\documentclass[final, 12pt]{elsarticle}

\usepackage{amssymb}
\usepackage{algorithmic}
\usepackage{algorithm}
\usepackage{soul}    
\usepackage{textcomp}
\usepackage{stfloats}
\usepackage{url}
\usepackage{verbatim}
\usepackage{graphicx}
\usepackage{color, xcolor}
\usepackage{verbatim}
\usepackage{float}
\usepackage{multirow}
\usepackage{booktabs}
\usepackage{amsmath}
\usepackage{hyperref} 
\newcommand\onedot{.}
\newcommand\eg{\emph{e.g}\onedot} 
\newcommand\ie{\emph{i.e}\onedot}

\journal{Expert Systems With Applications}

\begin{document}

\begingroup
\renewcommand{\thefootnote}{}
\footnotetext{\raggedright $^{*}$Corresponding author\\
Email addresses: zhaohaoru@sdpc.edu.cn (Haoru Zhao),
yufengwang@stu.ouc.edu.cn (Yufeng Wang), guzhaorui@ouc.edu.cn (Zhaorui Gu),
bingzh@ouc.edu.cn (Bing Zheng), zhenghaiyong@ouc.edu.cn (Haiyong Zheng)}
\addtocounter{footnote}{-1}
\endgroup

\begin{frontmatter}



\title{Context-Aware Mutual Learning for Blind Image Inpainting and Beyond}

\author[1]{Haoru~Zhao}
\author[2]{Yufeng~Wang}
\author[2]{Zhaorui~Gu}
\author[2,3]{Bing~Zheng}
\author[2]{Haiyong Zheng$^{*}$}
\address[1]{Department of Economic Crime Investigation, Shandong Police College, Jinan 250200, China}
\address[2]{College of Electronic Engineering, Ocean University of China, Qingdao 266000, China}
\address[3]{Sanya Oceanographic Institution, Ocean University of China, Sanya 572000, China}

\begin{abstract}
\small
Blind image inpainting, aiming to recover contaminated images in the case of unknown masks, is a challenging task. Motivated by the perspective of human vision and knowledge, blind image inpainting can be decomposed into two stages: mask estimation and image inpainting based on the estimated mask. The two-stage idea exhibits evident advantages in enhancing inpainting quality and augmenting the generalization capability of unknown real-world contamination by explicitly employing the estimated mask for image inpainting compared to one-stage scheme. This two-stage idea has also been intuitively implemented. However, existing two-stage methods excessively emphasize the unilateral relationship of mask estimation to image inpainting, and may overlook the mutual relations between them. Specifically, mask estimation can provide more contextual semantics for image inpainting to strengthen the understanding of semantics, and image inpainting can offer more contextual details (\eg, textures and edges) for mask estimation to improve the learning of details.
In this work, we propose a novel Context-Aware Mutual Learning (CAML) framework for blind image inpainting that joints mask estimation and image inpainting to mutually exploit contextual information. In the CAML framework, we design the Inpainting-Guided Context-Mutual (IGCM) learner to acquire the complementary contextual details from image inpainting for assisting mask estimation, and the Estimation-Guided Context-Mutual (EGCM) learner to strengthen the understanding of contextual semantics from mask estimation for assisting image inpainting.
Ablation studies validate the efficacy of our CAML. 
Extensive experiments show that our CAML achieves state-of-the-art performance on both blind image inpainting and additional vision tasks, \ie, snow removal, shadow removal, and watermark removal, indicating its superiority. Codes, more results, and details can be found at \href{https://github.com/zhenglab/CAML}{https://github.com/zhenglab/CAML}.

\end{abstract}



\begin{keyword}
Mutual Learning \sep
Context-Aware \sep
Blind Image Inpainting \sep
Snow Removal \sep
Shadow Removal \sep
Watermark Removal

\end{keyword}

\end{frontmatter}



\section{Introduction}
Image inpainting refers to recovering contaminated regions of an image that can be widely used in many applications, such as photo editing~\citep{barnes2009patchmatch,dekel2018sparse}, object removal~\citep{liu2018image,yu2019free,zhang2023adaptive}, and damaged image repairing~\citep{yan2018shift,zeng2019learning,wang2024road}. Recently, most image inpainting methods have achieved remarkable results based on an assumption that the mask of locating contaminated regions is known. This case can be called non-blind setting. However, in most practical applications, the masks are unavailable. Therefore, image inpainting with unknown masks, \ie, blind image inpainting, is more significant and useful for practical applications.

Compared to non-blind image inpainting, blind image inpainting is required to recover contaminated images in the case of unknown masks and thus is extremely more challenging. Specifically, due to the complexity of contamination (\eg, diverse contents with various shapes and positions) among different contaminated images, it is more difficult to effectively locate and recover contaminated regions without masks. From the perspective of human vision and knowledge, when recovering contaminated images with unknown masks, humans typically first recognize and locate contaminated regions by understanding the image context. Subsequently, they proceed to conceive and fill the content of contaminated regions by considering the pixel context. This process involves two stages: \emph{mask estimation} from the contaminated image and \emph{image inpainting} based on the estimated mask. The idea of this two-stage process was intuitively implemented by \citep{wang2020vcnet, wang2022ft}. Recently, one-stage solutions~\citep{zhao2022transcnn, phutke2023blind} were also proposed for blind image inpainting, which directly recovered the contents of contaminated regions from contaminated images.
Although the one-stage idea is more direct and simple, the lack of mask estimation may suffer from excessive interference of contaminated features that degrade the inpainting performance of the model. Additionally, these approaches may also easily remember the inherent pattern of contamination that influences the generalization of the model for unknown real-world contamination. Moreover, it is relatively unclear for the whole process of recovering contaminated images in the one-stage solution. In comparison, the two-stage idea is relatively complex and may require more computation costs. However, mask estimation enables accurate capture of non-contaminated features for inpainting contaminated regions to improve the quality of inpainting. It also adequately learns and understands the contextual differences between contaminated and non-contaminated regions to locate contamination, thereby enhancing its generalization ability to unknown real-world contamination. Furthermore, the two-stage design also provides a clearer explanation for the whole process of recovering contaminated images, which is more acceptable and understandable.

In the two stages of blind image inpainting, mask estimation is capable of recognizing multi-property contaminations in the contaminated images to generate masks. This makes mask estimation focus more on the understanding of contextual semantics in contaminated images. Image inpainting has the ability to fill contaminated regions with the assistance of the estimated masks for producing visually plausible images. As a result, image inpainting places more emphasis on learning realistic contextual details (\eg, textures and edges) in contaminated images. In this way, we argue that mask estimation and image inpainting are highly correlated, because they both essentially concentrate on the contextual information of contaminated images, although with different emphases.
Therefore, we consider that they can mutually cooperate to provide beneficial clues for each other. Specifically, image inpainting can offer more contextual details for mask estimation to improve the learning of details for achieving more accurate masks. Mask estimation can provide more contextual semantics for image inpainting to strengthen the understanding of semantics for facilitating more reasonable content. Existing two-stage solutions~\citep{wang2020vcnet, wang2022ft} fail to fully exploit the important mutual relations between mask estimation and image inpainting, as it overemphasizes the unilateral relationship of mask estimation to image inpainting. This may lead to unnatural and inaccurate details in mask estimation yet unreasonable semantics in image inpainting (refer to Figure~\ref{fig:QualitativeComparisonInpainting1} and Figure~\ref{fig:QualitativeComparisonInpainting2} for example). Hence, making full use of the correlation to strengthen the mutual cooperation between mask estimation and image inpainting, is very crucial for blind image inpainting.

\begin{figure}[t]
\centering
\includegraphics[width=\linewidth]{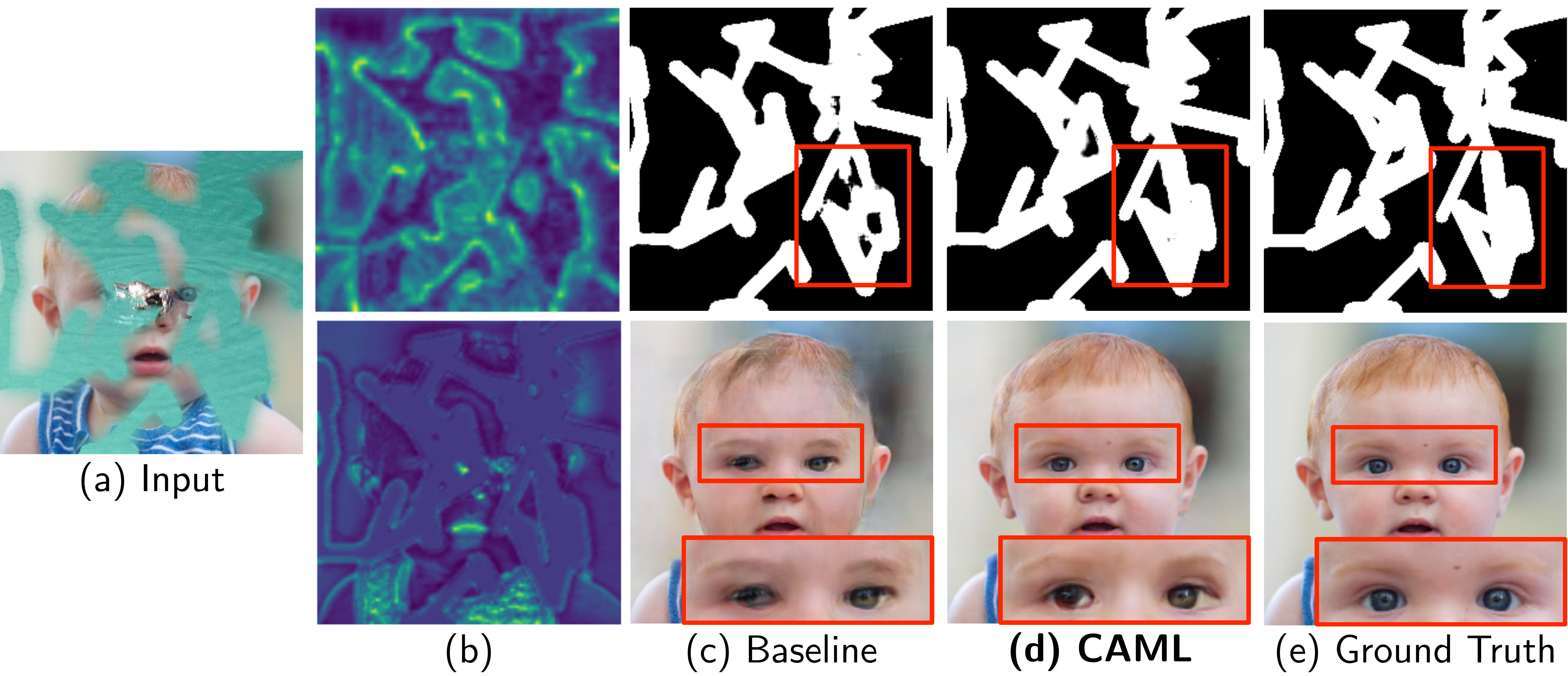}
\caption{Feature visualization and visual comparison on FFHQ (top: mask estimation, bottom: image inpainting). (a) The contaminated image (Input). (b) Visualized features after the last layer of both encoders in (c) Baseline, which comprises two basic encoder-decoder networks for mask estimation and image inpainting respectively without mutuation. (d) Our CAML leverages context-aware mutual learning of mask estimation and image inpainting to produce more accurate estimated mask (top) and visually realistic yet reasonable inpainted image (bottom). (e) Ground Truth.}
\label{fig:motivation}
\end{figure}

For better analysis, we employ two basic encoder-decoder networks for mask estimation and image inpainting respectively as the baseline of two-stage blind image inpainting. In baseline, we visualize features after the last layer of two encoders and results generated in these two networks, as shown in Figure~\ref{fig:motivation}. We discover that: (1) in mask estimation, the encoder feature is more abstract and focuses more on the main semantic differences between contaminated and non-contaminated regions (\eg, facial features and boundary features of different regions) rather than contextual details (see Figure~\ref{fig:motivation} (b) top for example). Due to the insufficient learning of contextual details, mask estimation produces an inaccurate result (see Figure~\ref{fig:motivation} (c) top for example); (2) In image inpainting, the encoder feature is vivid and detailed, which focuses more on contextual realistic details (\eg, textures and edges of clothes) compared with mask estimation. Due to the inadequate understanding of contextual semantics, image inpainting appears unreasonable and unrealistic semantic contents (note the eyes in Figure~\ref{fig:motivation} (c) bottom).

Motivated by the above discoveries, we decompose blind image inpainting into two mutual learning sub-tasks, \ie, mask estimation and image inpainting. We seek to deal with blind image inpainting via a mutual cooperation and joint learning framework between mask estimation and image inpainting. The framework can complement and benefit each other with mutual context information towards effective performance boost.
Correspondingly, we design an inpainting-guided learning strategy to transfer more complementary contextual details from image inpainting, for assisting mask estimation to learn finer details. We also devise an estimation-guided learning strategy to transfer and fuse more complementary contextual semantics, for helping image inpainting to strengthen the understanding of contextual semantics, thus yielding more reasonable and realistic inpainted results. Meanwhile, we consider absorbing valuable contextual information more mutually between mask estimation and image inpainting in an iterative manner.

Beyond blind image inpainting, we also explore the potential of our mutual cooperation and joint learning framework in other vision tasks. From the perspective of the object, blind image inpainting can be specialized as removing the objects from the images, thus we choose to apply our framework to snow removal, shadow removal, and watermark removal tasks as well as decompose these tasks into two stages of estimation and removing snow/shadow/watermark, demonstrating the powerful ability of our mutual learning framework with superior performance on both blind image inpainting and image object removal.

Our \textbf{contributions} include: (1) we propose a novel mutual cooperation and joint learning framework for blind image inpainting, named Context-Aware Mutual Learning (CAML); (2) we devise Inpainting-Guided Context-Mutual (IGCM) learner and estimation-Guided Context-Mutual (EGCM) learner for CAML to mutually learn complementary contextual information; (3) we conduct ablation studies to validate the efficacy of CAML, and extensive experiments on various datasets with complex contaminations to demonstrate that our CAML achieves state-of-the-art performance in blind image inpainting; (4) we further demonstrate the potential of our CAML in three object removal tasks, \ie, snow removal, shadow removal, and watermark removal, all producing very competitive results.

\section{Related Work}

\subsection{Non-Blind Image Inpainting}

Deep CNNs, especially generative adversarial networks (GANs)~\citep{goodfellow2014generative}, have shown remarkable capability in improving non-blind image inpainting performance. Since the pioneer method of using GAN to generate more visually plausible contents via directly predicting pixel values inside masks~\citep{pathak2016context}, plenty of GAN-based methods have been proposed for non-blind image inpainting. In order to handle irregular/free-form masks and exclude the distraction of invalid pixels, \citep{liu2018image} and \citep{yu2019free} respectively devised special Partial Convolution (PC) and Gated Convolution (GC) with mask updating strategies. 
Subsequently, \citep{zheng2019pluralistic} proposed a dual pipeline probabilistically principled framework (PICNet) that trade-offs between diversity and reconstruction. Several other methods~\citep{nazeri2019edgeconnect, guo2021image,wei2022ecnfp,shao2023two} exploited various types of intermediate representations as extra supervisions for non-blind image inpainting, such as texture, edge, and structure, among which, \citep{guo2021image} modeled the texture synthesis and structure reconstruction in a two-stream network (CTSDG) simultaneously.
Moreover, \citep{suvorov2022resolution} proposed a Fourier convolution-based network, a high receptive field perceptual loss, and large training masks for large mask inpainting (LaMa). \citep{li2022misf} addressed image inpainting as a filtering task and proposed multi-level interactive siamese filtering (MISF) to recover details and semantic information of contaminated regions at the semantic and image levels. \citep{liu2023coordfill} exploited the continuous implicit representation to propose a CoordFill framework via parameterized coordinate querying for high-resolution inpainting.
Recently, Transformer-based methods~\citep{dong2022incremental,liu2022reduce,zhou2023superior} have received excellent performance in image inpainting. For example, \citep{liu2022reduce} presented a patch-based auto-encoder and un-quantized Transformer for pluralistic inpainting.
Meanwhile, diffusion models~\cite{ho2020denoising}, praised for their powerful generative performance, have also been widely used in image inpainting. \citep{lugmayr2022repaint} employed an unconditional DDPM model as the generative prior (RePaint), incorporating information from the unmasked regions of the given image during the reverse diffusion process to guide generation. \citep{rombach2022high} performed the diffusion process in the latent space for efficiently generating inpainted images. \citep{yang2023uni} proposed a unified framework for multimodal image inpainting (Uni-paint) by utilizing a pre-trained diffusion model to perform inpainting across different modalities.
However, all these methods recovered the contaminated regions with the help of pre-defined masks. Different from existing non-blind inpainting methods, our work seeks to strengthen the contextual learning for blind image inpainting via the mutual learning of mask estimation and image inpainting based on the estimated mask.

\subsection{Blind Image Inpainting}

Previous methods~\citep{dong2012wavelet,yan2013restoration,cai2017blind,liu2019deep} for blind image inpainting usually assumed that contaminated regions are filled with simple data distributions, such as constant values or noises. This simple setting makes these methods only suitable for images contaminated by simple data distribution, but it could be invalid when the contamination is complex with various properties, \eg, graffiti, text, and even image. Recent studies~\citep{wang2020vcnet, wang2022ft, zhao2022transcnn, wang2023self, phutke2023blind} relaxed this assumption and employed images with complex data distributions as contaminations for blind inpainting. These studies can be divided into one-stage and two-stage methods. In the one-stage methods, \citep{zhao2022transcnn} merged the mask estimation and image inpainting to present a one-stage Transformer-CNN hybrid autoencoder (TransCNN-HAE) solution for blind image inpainting. \citep{phutke2023blind} proposed a one-stage transformer-based architecture (OmniWavNet) consisting of a wavelet query multi-head attention transformer block and omni-dimensional gated attention for blind inpainting.
In the two-stage methods, \citep{wang2020vcnet} proposed a two-stage visual consistency network (VCN) for addressing the problem with two sub-tasks of mask estimation and image inpainting. \citep{wang2022ft} proposed a two-stage method that consists of a Transformer-based mask detection module to estimate mask and a top-down refinement module to inpaint image. \citep{wang2023self} also divided blind inpainting into two critical problems, \ie, ``where to inpaint'' and ``how to inpaint''. ``Where to inpaint'' aims to estimate the location of corruption (mask). ``How to inpaint'' aims to image inpainting based on the estimated mask, and in order to synthesize more realistic texture, they additionally predict a layout map for image inpainting. 
Compared with the one-stage ideas, the two-stage approaches can improve the quality of inpainting, exhibit stronger generalization for unknown real-world contamination, and provide a clearer understanding of the whole inpainting process. Therefore, in our work, we also adopt a two-stage solution for blind inpainting and use various contaminations with complex data distributions. Differently, we devote to investigating context-aware mutual learning between two sub-tasks of mask estimation and image inpainting, for learning complementary contexts to benefit each other for boosting the performance of blind inpainting.

\subsection{Mutual Learning}

Mutual learning~\citep{zhang2018deep} aims at making a group of students to learn collaboratively and teach each other for solving tasks, which has shined in many CV tasks recently, such as object detection~\citep{wu2019mutual,zhai2021mutual}, person re-identification~\citep{hong2021fine}, human parsing and pose estimation~\citep{nie2018mutual,he2020grapy}, and spatial-temporal video super-resolution~\citep{hu2022spatial}. Among them, \citep{wu2019mutual} utilized mutual learning to interweave saliency detection and foreground contour detection, and the network can benefit from the strengths of both tasks, leading to more entirely highlighted salient regions. \citep{hong2021fine} proposed a dense interactive mutual learning framework (FSAM) to transfer knowledge between appearance stream and shape stream to complement body shape features in appearance features for cloth-changing person re-identification. \citep{nie2018mutual} presented mutual learning to adapt model (MuLA) for joining human parsing and pose estimation to boost their performance simultaneously.
Our work also contributes to the study of mutual learning, diving deep into mutual learning for blind inpainting and beyond, providing the context-aware mutual learning framework for a series of object removal tasks.

\begin{figure*}[!t]
\centering
\includegraphics[width=\linewidth]{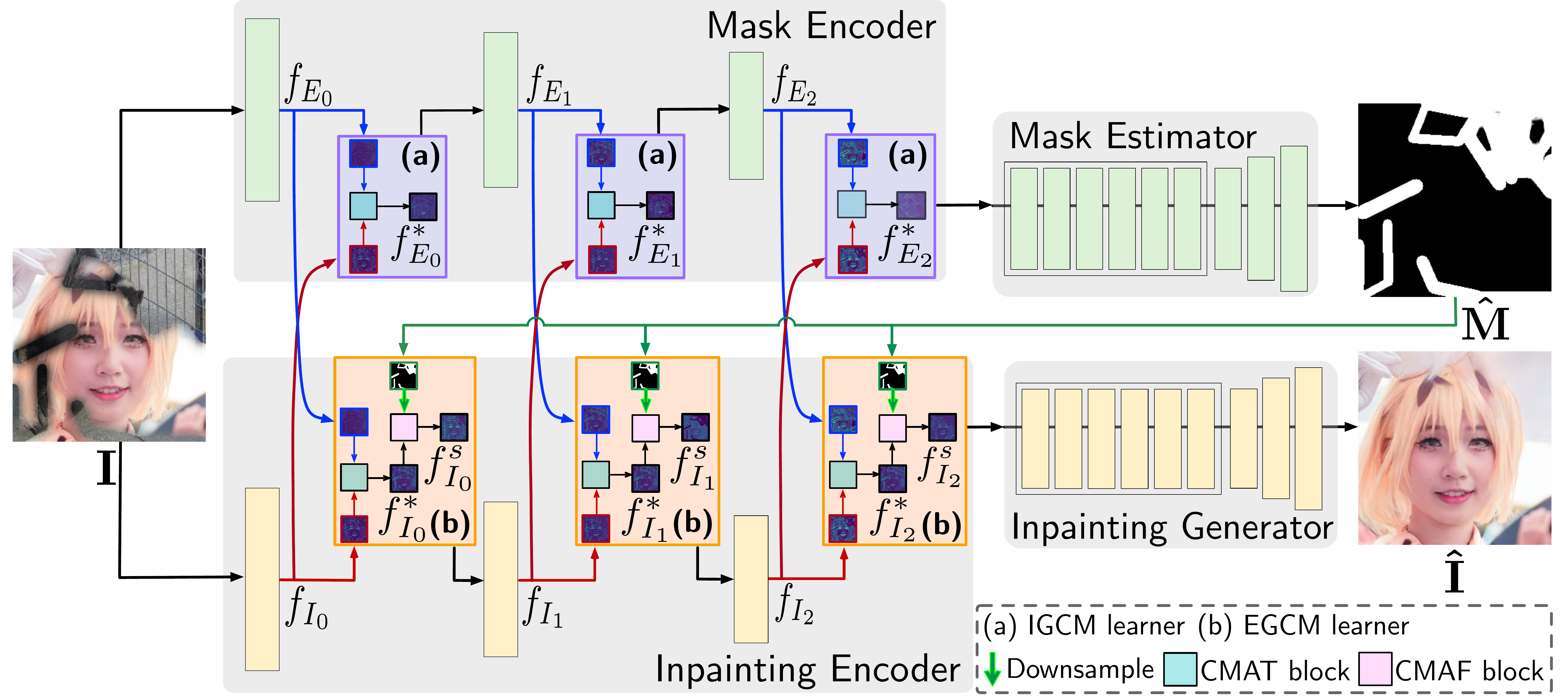}
\caption{Our CAML framework contains two core components (a) Inpainting-Guided Context-Mutual (IGCM) learner and (b) Estimation-Guided Context-Mutual (EGCM) learner.
Context-Mutual Adaptive Transfer (CMAT) block and Context-Mutual Adaptive Fusion (CMAF) block are core operations designed for IGCM learner and EGCM learner, with details shown in Figure~\ref{fig:CMAT} and Figure~\ref{fig:CMAF} respectively. Top: mask estimation. Bottom: image inpainting.}
\label{fig:framework}
\end{figure*}

\section{Context-Aware Mutual Learning}

In this section, we introduce the proposed Context-Aware Mutual Learning (CAML) framework for blind image inpainting, as shown in Figure~\ref{fig:framework}. The CAML framework consists of two mutual learning sub-tasks: mask estimation and image inpainting. Mask estimation contains a mask encoder to extract contextual estimated information and a mask estimator to estimate mask. Image inpainting includes an inpainting encoder to extract contextual inpainted information, an inpainting generator to produce the inpainted image, as well as a global discriminator and a contaminated discriminator (the discriminators are not shown in Figure~\ref{fig:framework}) to distinguish between the global result and the contaminated region from the corresponding reals respectively. 

Specifically, in the CAML framework, given a contaminated image $\mathbf{I}\in \mathbb{R}^{h\times w\times 3}$ for mask estimation and image inpainting. Mask estimation aims to estimate the mask $\mathbf{\hat{M}} \in {[0,1]}^{h\times w\times 1}$ for locating contaminated regions through mask encoder and mask estimator. Image inpainting is devoted to yielding the image $\mathbf{\hat{I}}\in \mathbb{R}^{h\times w\times 3}$ with visually reasonable and realistic appearance through inpainting encoder and inpainting generator. Mathematically, the contaminated image $\mathbf{I}$ in the setting of blind image inpainting can be formulated as:
\begin{equation}
\mathbf{I}=\mathbf{I}_{gt}\odot(1-\mathbf{M})+\mathbf{N}\odot\mathbf{M},
\label{eq:contamination}
\end{equation}
where $\mathbf{I}_{gt}\in \mathbb{R}^{h\times w\times 3}$ is the ground truth image without contamination, $\mathbf{M}\in \mathbb{R}^{h\times w\times 1}$ is the binary ground truth mask ($1$ and $0$ represent contaminated and non-contaminated pixel respectively), $\mathbf{N}\in \mathbb{R}^{h\times w\times 3}$ is a noisy visual signal, and $\odot$ denotes element-wise product. Similar to~\citep{wang2020vcnet}, we adopt complex images or constant values as $\mathbf{N}$ for training. Besides, during the training process, we also randomly perform iterative Gaussian smoothing~\citep{wang2018image} on the contaminated region and employ alpha blending in the contact regions between $\mathbf{I}_{gt}$ and $\mathbf{N}$, making the contamination more indistinguishable.

In order to transfer more complementary contextual details (\eg, textures and edges) from image inpainting for helping mask estimation and more complementary contextual semantics from mask estimation for assisting image inpainting, we devise the Inpainting-Guided Context-Mutual (IGCM) learner and estimation-Guided Context-Mutual (EGCM) learner respectively as the core components of our CAML framework. Notably, the IGCM learner and the EGCM learner can be embedded into any layer of the network for mutual learning. Here, we equip the IGCM learner and the EGCM learner into each layer of mask encoder and inpainting encoder to adequately capture complementary contextual information for the following mask estimator and inpainting generator respectively. We also conduct the empirical study in \S\ref{sec:experiment:CAML} for reference. In the following, we will discuss the details of the IGCM learner in \S\ref{sec:method:IGCM} and the EGCM learner in \S\ref{sec:method:EGCM}. Besides, we will introduce the loss functions for CAML framework in \S\ref{sec:method:Loss} and some implementation details in \S\ref{sec:method:details}.


\subsection{Inpainting-Guided Context-Mutual (IGCM) Learner}
\label{sec:method:IGCM}

\begin{figure}[t]
\centering
\includegraphics[width=\linewidth]{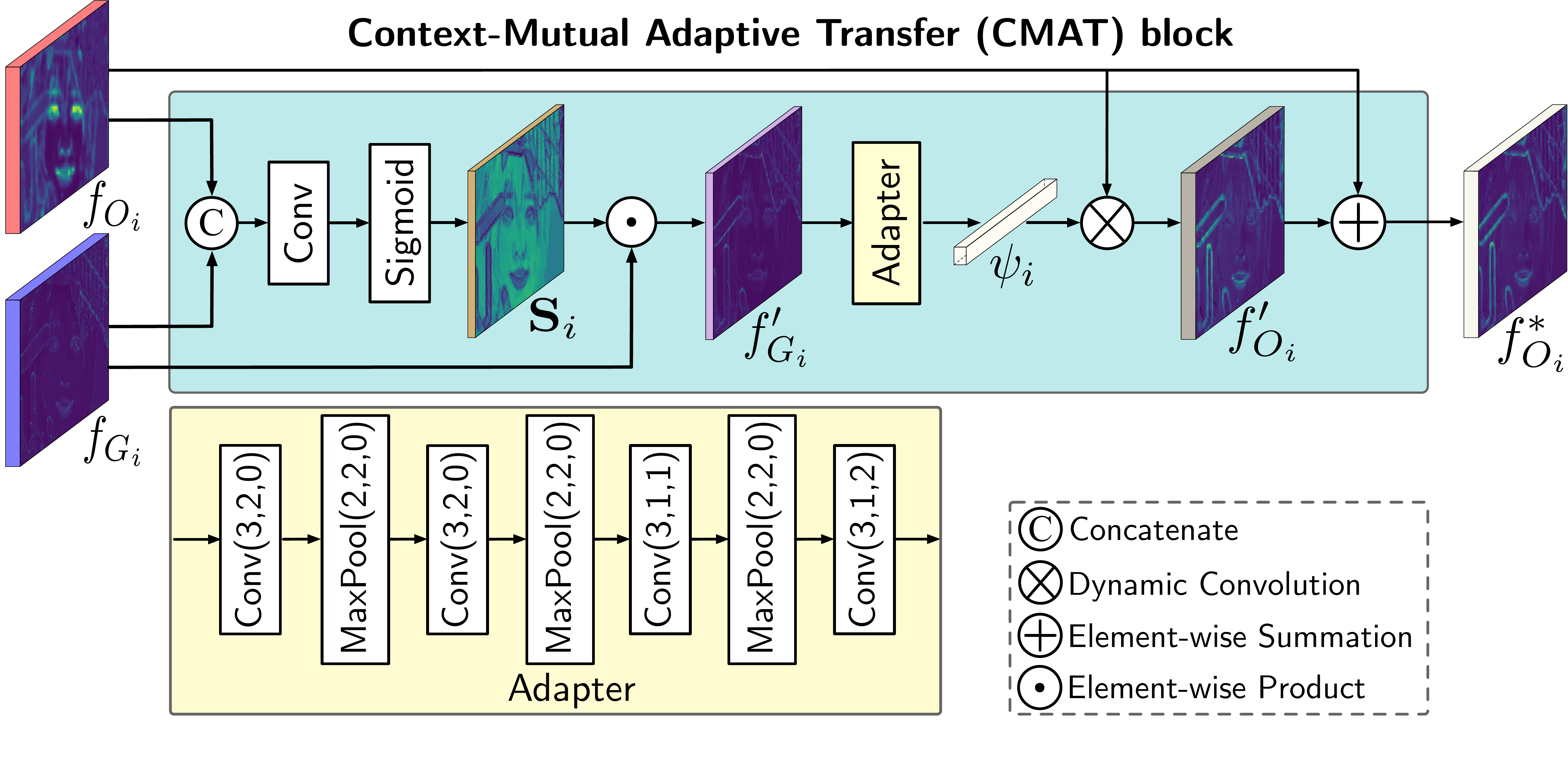}
\caption{Details of Context-Mutual Adaptive Transfer (CMAT) block. Adapter is a critical component of CMAT, where ``Conv($k$, $s$, $p$)'' denotes convolution with kernel size $k$, stride $s$, and padding $p$; ``MaxPool($k$, $s$, $p$)'' denotes max pooling with kernel size $k$, stride $s$, and padding $p$.}
\label{fig:CMAT}
\end{figure}

For IGCM learner, we seek to learn mutual relations between mask estimation and image inpainting for capturing more complementary contextual details (\eg, textures and edges) from image inpainting to assist mask estimation. IGCM learner is embedded into each layer of the mask encoder, and the structure is illustrated in Figure~\ref{fig:framework} (a). Specifically, IGCM learner receives contextual features $f_{E_i}$ (where $i=0, 1, 2$) from the mask encoder of mask estimation and $f_{I_i}$ from the inpainting encoder of image inpainting, and outputs the transferred feature $f^*_{E_i}$ with more contextual details via Context-Mutual Adaptive Transfer (CMAT) block. 

\textbf{Context-Mutual Adaptive Transfer (CMAT) Block.} 
CMAT block is devised to discover valuable complementary contextual clues by adaptively learning dynamic context-aware parameters from the guiding task. Subsequently, CMAT block precisely transfers these clues to the original task to enhance its representation. Figure~\ref{fig:CMAT} illustrates the CMAT block, it receives the original features $f_{O_i}$ and the guiding features $f_{G_i}$ from the original task and guiding task respectively as inputs, and outputs the transferred features $f^{*}_{O_i}$.

Specifically, in order to fully explore more valuable complementary contextual clues, we first try to learn a soft gating to control the selection of the guiding feature for eliminating the excessive interference of some unbeneficial features in the guiding feature to the original feature. In this process, we learn the soft gatings $\mathbf{S}_i$ by adaptively concatenated $f_{O_i}$ and $f_{G_i}$ as follows:
\begin{equation}
\mathbf{S}_i = {\rm Sigmoid}({\rm Conv}({\rm Concat}(f_{O_i}, f_{G_i})))),
\end{equation}
where $\rm Concat(\boldsymbol{\cdot})$ is channel-wise concatenation, $\rm Conv(\boldsymbol{\cdot})$ is $1 \times 1$ convolution, and ${\rm Sigmoid}(\boldsymbol{\cdot})$ is sigmoid activation. We then employ the soft gatings $\mathbf{S}_i$ for the guiding features $f_{G_i}$, yielding the selected guiding features $f'_{G_i}$:

\begin{equation}
f'_{G_i} =\mathbf{S}_i \odot f_{G_i},
\end{equation}
where $\odot$ is the element-wise product. 

Considering that the selected guiding features $f'_{G_i}$ hide more valuable complementary contextual clues, we intend to adaptively mine these cues by learning the context-aware parameters. Specifically, we first acquire the context-aware parameters $\psi_i$ via the learnable adapter $\rm Adapter(\boldsymbol{\cdot})$:

\begin{equation}
\psi_i = {\rm Adapter}(f'_{G_i}), 
\end{equation}
where the adapter $\rm Adapter(\boldsymbol{\cdot})$ consists of 4 layers of $3 \times 3$ convolution and 3 layers of $2 \times 2$ max pooling. The adapter extracts and compresses the guiding features through convolution and pooling operations to generate dynamic context-aware parameters, which are critical to helping CMAT discover valuable complementary context clues for effectively enhancing the feature representation of the original task. We then replace the static convolution kernel with the dynamic context-aware parameters $\psi_i$ and conduct the dynamic convolution operation on the original features $f_{O_i}$. Through dynamic convolution, $\psi_i$ gradually involves useful complementary contextual clues from the selected guiding features $f'_{G_i}$. As a result, we obtain the enhanced original features $f^{'}_{O_i}$ with more complementary contextual. The process can be formulated as:
\begin{equation}
f^{'}_{O_i} = \psi_i \otimes f_{O_i},
\end{equation}
where $\otimes$ denotes dynamic convolution with the dynamic context-aware parameters $\psi_i$.
We finally transfer the enhanced original features $f'_{O_i}$ to original features $f_{O_i}$ via element-wise summation, yielding the transferred features:
\begin{equation}
f^{*}_{O_i} = f'_{O_i} + f_{O_i}.
\end{equation}

Our design of CMAT block is inspired by~\citep{wu2019mutual}, but differently, it lacks soft gatings $\mathbf{S}_i$ for the guiding features $f_{G_i}$ and acquires the dynamic parameters directly from the $f_{G_i}$. In our study, we adopt the selected guiding feature to obtain the dynamic context-aware parameters so that these parameters contain more useful complementary contextual information. This information can be applicable to the original task for further enhancing the representation of the original task. 

In CMAT block of IGCM learner, we regard mask estimation as an original task and image inpainting as a guiding task. The contextual features $f_{E_i}$ obtained from the mask estimation are regarded as the original features $f_{O_i}$, while the contextual features $f_{I_i}$ obtained from the image inpainting are regarded as the guiding features $f_{G_i}$. The transferred features of CMAT block in IGCM learner can be represented as $f^*_{E_i}$ with more complementary contextual details (\eg, textures and edges), as shown in Figure~\ref{fig:framework} (a).

\subsection{Estimation-Guided Context-Mutual (EGCM) Learner}
\label{sec:method:EGCM}

For leveraging the guidance of mask estimation to obtain complementary contextual semantics by learning the mutual relations between mask estimation and image inpainting, we design EGCM learner for image inpainting as illustrated in Figure~\ref{fig:framework} (b). 
EGCM learner is embedded into each layer of the inpainting encoder. It receives the contextual features $f_{I_i}$ from the inpainting encoder of image inpainting, $f_{E_i}$ from the mask encoder of mask estimation, and the estimated mask from mask estimation as input. It outputs context-mutual fused features $ f^s_{I_i} $ with more valid contextual information via CMAT block and then Context-Mutual Adaptive Fusion (CMAF) block. 

In EGCM learner, CMAT block is used to extract more complementary context semantics in mask estimation, which are subsequently transferred to enhance the context features in image inpainting. Therefore, the contextual features in image inpainting encompass richer contextual semantic features. Among them, we regard image inpainting as an original task and mask estimation as a guiding task. The contextual features $f_{I_i}$ obtained from the image inpainting are regarded as the original features $f_{O_i}$, while the contextual features $f_{E_i}$ obtained from the mask estimation are regarded as the guiding features $f_{G_i}$. The transferred features of CMAT
block in EGCM learner can be represented as $f^*_{I_i}$ with more complementary contextual semantics, as shown in Figure~\ref{fig:framework} (b).

\begin{figure}[t]
\centering
\includegraphics[width=\linewidth]{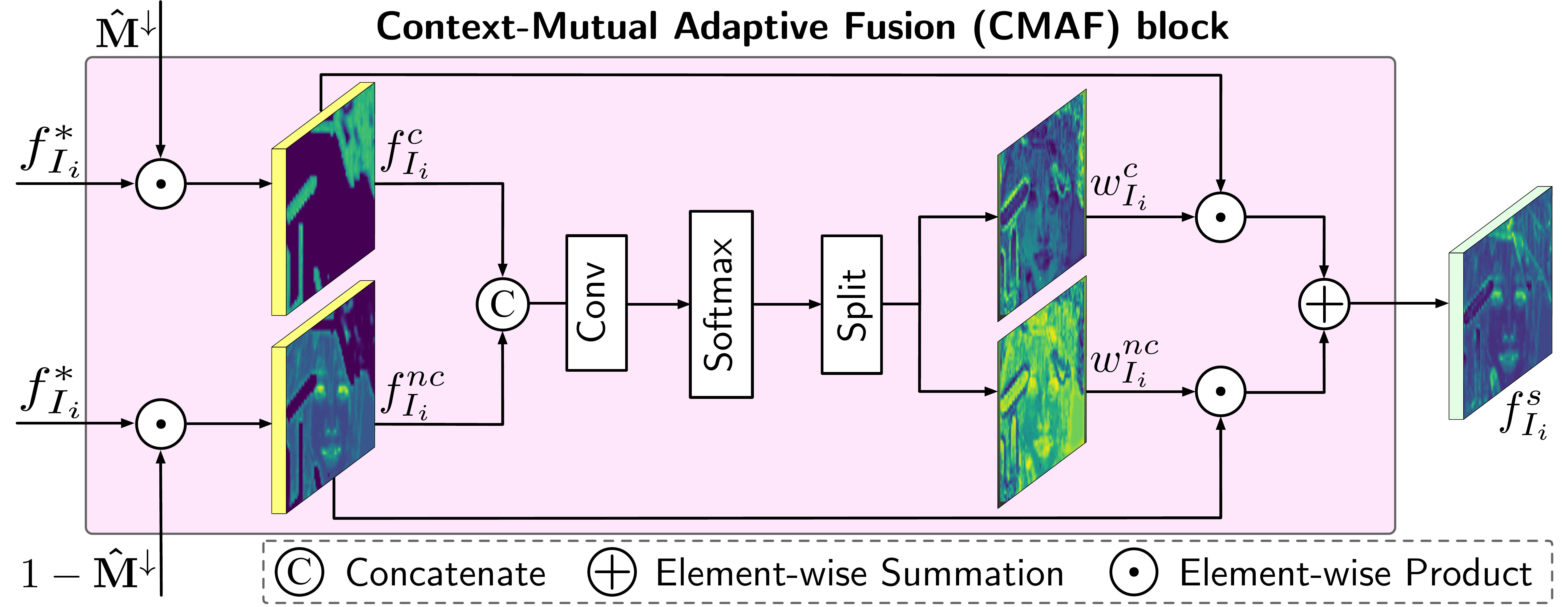}
\caption{Details of Context-Mutual Adaptive Fusion (CMAF) block.}
\label{fig:CMAF}
\end{figure}

\textbf{Context-Mutual Adaptive Fusion (CMAF) Block.} 
CMAF block aims at adaptively fusing contaminated and non-contaminated features in the transferred features to obtain valid contextual information for further image inpainting, as shown in Figure~\ref{fig:CMAF}. The contaminated features $f^c_{I_i}$ and non-contaminated features $f^{nc}_{I_i}$ are separated from the transferred features $f^*_{I_i}$ via:
\begin{align}
 f^c_{I_i} = f^*_{I_i} \odot \mathbf{\hat{M}^{\downarrow}}, 
\quad f^{nc}_{I_i} = f^*_{I_i} \odot (1 - \mathbf{\hat{M}^{\downarrow}}),
\end{align}
where $\mathbf{\hat{M}^{\downarrow}}$ is the mask obtained by downsampling estimated mask $\mathbf{\hat{M}}$ for corresponding the spatial size of $f^*_{O_i}$, and $\odot$ is element-wise product. Considering that both contaminated and non-contaminated features contain more or less valid and invalid contextual features for image inpainting, we seek to learn the pixel-wise weights $w^c_{I_i}$ and $w^{nc}_{I_i}$ for adaptively determining how many the valid features of contaminated and non-contaminated regions should be fused by:
\begin{align}
& w^c_{I_i}, w^{nc}_{I_i} = {\rm Split}({\rm Softmax}({\rm Conv}({\rm Concat}(f^c_{I_i}, f^{nc}_{I_i})))),
\end{align}
where $\rm Concat(\boldsymbol{\cdot})$ is channel-wise concatenation, $\rm Conv(\boldsymbol{\cdot})$ is $1 \times 1$ convolution, $\rm Softmax(\boldsymbol{\cdot})$ is channel-wise softmax, and $\rm Split(\boldsymbol{\cdot})$ means channel-wise split operation, that is, the input is evenly split into two branches at the channel level. Finally, we adaptively fuse $f^c_{I_i}$ and $f^{nc}_{I_i}$ to yield the context-mutual fused feature $f^s_{I_i}$: 
\begin{equation}
f^s_{I_i}= f^c_{I_i} \odot w^c_{I_i} + f^{nc}_{I_i} \odot w^{nc}_{I_i}.
\end{equation}

By performing the adaptive weighted fusion of features from the contaminated and non-contaminated regions, the features that are beneficial for inpainting are highlighted, while reducing the negative impact of mask estimation errors on the repair quality. Ultimately, the fused features become more accurate, allowing for effective utilization of useful information from the contaminated regions, thereby significantly improving the overall inpainting performance.

Actually, CMAF block serves as a critical component in EGCM learner. The introduction of pixel-level weights $w^c_{I_i}$ and $w^{nc}_{I_i}$ not only alleviates the performance degradation of inpainting caused by the error of mask estimation, but also adaptively extracts more valid contextual information from the contaminated and non-contaminated regions to $f^s_{I_i}$. Consequently, CMAF block successfully captures the valid contextual information to produce more reasonable content.

\subsection{Loss Functions}
\label{sec:method:Loss}

We employ binary cross-entropy loss for mask estimation task, as well as adversarial loss, reconstruction loss, perceptual loss, and style loss for image inpainting task.

\textbf{Adversarial Loss.} 
We adopt a global discriminator $D_G$ and a contaminated discriminator $D_C$ to distinguish the global region $\mathbf{\hat{I}}$ and the contaminated region $\mathbf{\hat{I}}^c$ from the corresponding real ones $\mathbf{I}_{gt}$ and $\mathbf{I}^c_{gt}$ respectively, defined as follows: 
\begin{align}
{\mathcal L}^{G}_{adv}(F_{II},D_G)=&{\mathbb{E}}_{\mathbf{I}_{gt}}[\log (D_G(\mathbf{I}_{gt}))] \nonumber \\
& + \mathbb{E}_{\mathbf{\hat{I}}}[\log (1-D_G(\mathbf{\hat{I}}))], \\
 {\mathcal L}^{C}_{adv}(F_{II},D_C)=&{\mathbb{E}}_{\mathbf{I}^c_{gt}}[\log (D_C(\mathbf{I}^c_{gt}))] \nonumber \\
& + \mathbb{E}_{\mathbf{\hat{I}}^c}[\log (1-D_C(\mathbf{\hat{I}}^c))],
\end{align} 
where $\mathbf{I}^c_{gt}=\mathbf{I}_{gt}\odot\mathbf{M}$, $\mathbf{\hat{I}}^c=\mathbf{\hat{I}}\odot\mathbf{M}$, $F_{II}$ is the image inpainting function, which is trained via the minimax game.
And our final adversarial loss is:
\begin{equation}
\mathcal L_{adv}=\left({\mathcal L}^{G}_{adv} + {\mathcal L}^{C}_{adv}\right)/{2}.
\end{equation}   

\textbf{Total Loss.}
Our total loss functions ${\mathcal L}_{F_{ME}}$ and ${\mathcal L}_{F_{II}}$ for mask estimation and image inpainting respectively, are as follows:
\begin{align}
{\mathcal L_{F_{ME}}}=&\mathcal L_{BCE}, \\
{\mathcal L_{F_{II}}}=&\lambda_{adv}\mathcal L_{adv} + \lambda_{rec}\mathcal L_{rec} + \lambda_{perc}\mathcal L_{perc} 
+ \lambda_{style}\mathcal L_{style},
\end{align}
where $F_{ME}$ is the mask estimation function, $\mathcal L_{BCE}$ is binary cross-entropy loss between $\mathbf{\hat{M}}$ and $\mathbf{M}$, $\mathcal L_{rec}$ adopts $\ell_1$ distance between $\mathbf{\hat{I}}$ and $\mathbf{I}_{gt}$, $\mathcal L_{perc}$ and $\mathcal L_{style}$ denote perceptual~\citep{johnson2016perceptual} and style~\citep{gatys2016image} losses respectively, and the $\lambda$s are weights to balance different losses.

\subsection{Implementation Details}
\label{sec:method:details}

\textbf{Detailed settings of our CAML.}
In our CAML framework, we utilize two fundamental encoder-decoder structures as the baseline: one for mask estimation and the other for image inpainting. The mask estimation comprises a mask encoder and a mask estimator, while the image inpainting includes an inpainting encoder and an inpainting generator. Building on this baseline, CAML introduces three IGCM learners and three EGCI learners to extract complementary contextual information from both image inpainting and mask estimation.
Furthermore, inspired by MUSICAL~\cite{wang2019musical}, we implemented a DenseNet-like structure~\cite{huang2017densely} as our global discriminator, and a multi-scale PatchGAN~\cite{isola2017image} as the contamination discriminator. Detailed configurations of the mask encoder, mask estimator, inpainting encoder, and inpainting generator in the CAML framework as shown in Table~\ref{tab:detailed}.
Furthermore, inspired by MUSICAL~\cite{wang2019musical}, we implemented a DenseNet-like structure~\cite{huang2017densely} as our global discriminator, and a multi-scale PatchGAN~\cite{isola2017image} as the contamination discriminator. Detailed configurations of the mask encoder, mask estimator, inpainting encoder, and inpainting generator in the CAML framework as shown in Table~\ref{tab:detailed}.
\begin{table}[ht]
\footnotesize
\centering
 	\setlength{\tabcolsep}{0.8mm}
	\begin{tabular}{lclc}
		\hline
	 Estimating Encoder & Output  & Inpainting Encoder & Output \\
		\hline
 ResBlock(3, 64, 3, 1, 1) &  256  & Conv(3, 64, 7, 1, 3)+IN(64)+ReLU() &  256\\
IGCM learner &  256 & EGCM learner & 256 \\
  ResBlock(64, 128, 3, 2, 1) &  128 & Conv(64, 128, 3, 2, 1)+IN(128)+ReLU() & 128\\
 IGCM learner &  128 & EGCM learner & 128\\
 ResBlock(128, 256, 3, 2, 1) &  64 & Conv(128, 256, 3, 2, 1)+IN(256)+ReLU() & 64\\
 IGCM learner &  64 & EGCM learner & 64 \\
		\hline
Mask Estimator & Output  & Inpainting Generator & Output  \\
\hline
 ResBlock(256, 256, 3, 1, 1) $\times 6$ & 64 & ResBlock(256, 256, 3, 1, 1) $\times 6$ & 64  \\
 Upsample(2) &  128 & Upsample(2) &  128 \\
 ResBlock(256, 128, 3, 1, 1) &  128  & Conv(256, 128, 5, 1, 2)+IN(128)+ReLU() & 128\\
 Upsample(2) &  256 & Upsample(2) &  256\\
 ResBlock(128, 64, 3, 1, 1) &  256 & Conv(128, 64, 5, 1, 2)+IN(128)+ReLU() & 256\\
 Conv(64, 1, 3, 1, 1)$\rightarrow$ output& 256 &  Conv(64, 3, 7, 1, 3)$\rightarrow$ output& 256\\
  		\hline


	\end{tabular}
	\caption{Detailed settings of our estimating encoder, mask estimator, inpainting encoder, and inpainting generator. ``ResBlock($c_{in}$, $c_{out}$, $k$, $s$, $p$)'' denotes residual block~\cite{he2016deep} with $c_{in}$ input channels, $c_{out}$ output channels, kernel size $k$,
stride $s$, and padding $p$. ``Conv($c_{in}$, $c_{out}$, $k$, $s$, $p$)'' denotes convolution with $c_{in}$ input channels, $c_{out}$ output channels, kernel size $k$, stride $s$, and padding $p$. ``Upsample($s$)'' means nearest-neighbor upsampling with a scale factor of $s$. ``IN($n$)'' means instance normalization~\cite{ulyanov2017improved} with $n$ dimensions. ``Output'' means output size.}
	\label{tab:detailed}
\end{table}

\textbf{Iterative Training Procedure.} Notably, our CAML framework is trained end-to-end, with the two sub-tasks updated alternately. To fully exploit the mutual benefits between mask estimation and image inpainting, we employ an iterative training strategy. Specifically, for each iteration, we first train mask estimation to obtain the estimated mask $\mathbf{\hat{M}}$ with the help of the contextual features $f_{I_i}$ ($i=0, 1, 2$) of image inpainting. Then, we train image inpainting relying on the contextual features $f_{E_i}$ of image estimation and estimated mask $\mathbf{\hat{M}}$ to predict the inpainted image. 
The main consideration for this iterative training strategy is that in blind image inpainting tasks, the accuracy of mask estimation is critical to the quality of subsequent image inpainting. By accurately estimating the mask first, we can provide more precise guidance for image inpainting, allowing the model to make fuller use of information from non-contaminated regions to repair the contaminated regions, thereby effectively enhancing the quality and consistency of the inpainted images. Meanwhile, the iterative training strategy helps avoid interference in the early stages of image inpainting caused by inaccurate mask estimation, enabling the model to better learn suitable feature representations. We perform our CAML framework by receiving only the contaminated image as the input for blind image inpainting. At the inference phase, our CAML framework still performs mask estimation to estimate mask, and then performs image inpainting based on the estimated mask in one forward pass. The training procedure is listed in Algorithm~\ref{alg1}.

\begin{algorithm}[t!]
	\caption{Training Procedure of CAML.}
	\label{alg1}
	\begin{algorithmic}[1]
	\FOR{number of training iterations}
	\STATE Sample contaminated image $\mathbf{I}$ randomly from the training data;
 \STATE \textbf{Training the mask estimation:}
	    \STATE Get the contextual feature $f_{I_i}$ by inpainting encoder without EGCM learners;
	    \STATE Get the estimated mask $\mathbf{\hat{M}}=F_{EM}(\mathbf{I},f_{I_i})$;
	    \STATE Update mask estimation by $\mathcal L_{F_{EM}}$;
      \STATE \textbf{Training the image inpainting:}
	    \STATE Get the contextual features $f_{E_{i}}$ by mask encoder without IGCM learners;
	    \STATE  Get the inpainted image $\mathbf{\hat{I}}=F_{II}(\mathbf{I}, \mathbf{\hat{M}}, f_{E_{i}})$;
	    \STATE Update image inpainting by $\mathcal L_{F_{II}}$;
	\ENDFOR
	\end{algorithmic}  
\end{algorithm}

\textbf{Other Settings.}  
The model is implemented in PyTorch, and the training is launched on a single RTX 3090 GPU with the batch size of 4. We train our model using Adam optimizer~\citep{kingma2014adam} with parameters of $\beta_1=0.5$, $\beta_2=0.999$, learning rate $\alpha_1=0.00005$ for mask estimation, learning rate $\alpha_2=0.0001$ for image inpainting, and learning rate $\alpha_3=0.0002$ for global and contaminated discriminators. All the images and corresponding masks are resized to $256\times256$ resolution. The hyper-parameters of the loss function are empirically set to $\lambda_{adv}=0.1$, $\lambda_{rec}=10$, $\lambda_{perc}=10$, and $\lambda_{style}=250$ in our experiments. 


\section{Experiments on Blind Image Inpainting}

\subsection{Experimental Setup}

\textbf{Datasets.}
We evaluate the performance of our CAML for blind image inpainting on five public image datasets, considering three cases of faces (FFHQ~\citep{karras2019style}, CelebA-HQ~\citep{karras2018progressive}), objects (ImageNet~\citep{deng2009imagenet}), and scenes (Paris StreetView~\citep{doersch2012makes}, Places2~\citep{zhou2017places}), described as follows.
\begin{itemize}
\item FFHQ contains $70,000$ high-quality human face images, including $60,000$ training images and $10,000$ testing images. We use the default official training images for our training, and randomly select $2,000$ official testing images for our testing.
\item CelebA-HQ contains $30,000$ aligned human face images. We randomly select $28,000$ images for our training, and keep the others for our testing.
\item ImageNet is a large-scale object dataset and contains $1,000$ object classes with more than $1.4$ million images. We use the default official training images for our training, and randomly select $4,000$ official testing images for our testing.
\item Paris StreetView contains $15,000$ street view images in Paris, including $14,900$ training images and $100$ testing images. We keep the default official split for our training and testing.
\item Places2 contains more than $18$ million natural images from $365$ scene categories. We use the default official training set for our training and randomly select $1,000$ images of $10$ scene categories from the official testing set for our testing.
\end{itemize}

Following~\citep{wang2020vcnet,zhao2022transcnn}, we set the contaminated pattern (noisy visual signal $\mathbf{{N}}$ in Eq.~\ref{eq:contamination}) and the mask (binary mask $\mathbf{M}$ in Eq.~\ref{eq:contamination}) as follows. For FFHQ, CelebA-HQ, Paris StreetView, and Places2, we randomly adopt constant values and images from ImageNet as contaminated patterns for training and testing. For ImageNet, we randomly adopt constant values and images from Places2 as contaminated patterns for training and testing. Meanwhile, we use the free-form mask~\citep{yu2019free} with various mask ratios as the mask for training and testing. Further, we also employ graffiti (the contaminated ratio is 10-20\%), text (the contaminated ratio is 0-20\%), and even random image occlusion (the contaminated ratio is 0-20\%) from unseen Flowers~\citep{nilsback2008automated} and Stanford Cars~\citep{krause20133d} as unseen contaminated patterns for evaluation. Moreover, we conduct ablation studies to validate the efficacy of our CAML on FFHQ dataset with random mask ratios as the binary mask $\mathbf{M}$.

\begin{table*}[t!]
 \centering
\scriptsize
\setlength{\tabcolsep}{1mm}
	\begin{tabular}{c c c c c c c c c c c c c}
			\hline
			\multicolumn{1}{c}{\multirow{2}{*}{Metric}} &
			\multicolumn{1}{c}{\multirow{2}{*}{Method}} &
			\multicolumn{3}{c}{\multirow{1}{*}{FFHQ}} & &
			\multicolumn{3}{c}{\multirow{1}{*}{CelebA-HQ}} & &
          
			\multicolumn{3}{c}{\multirow{1}{*}{ImageNet}}
			\\
   \cline{3-5}  \cline{7-9}  \cline{11-13}
   & & $0$-$20\%$ & $20$-$40\%$ & $40$-$60\%$ && $0$-$20\%$ & $20$-$40\%$ & $40$-$60\%$ && $0$-$20\%$ & $20$-$40\%$ & $40$-$60\%$ \\
   
		   	\hline
		   	\multirow{2}{*}{BCE$\downarrow$}
    & VCN & $0.79$ & $0.69$ & $0.87$ && $0.66$ & $0.82$ & $0.92$ && $0.66$ & $0.81$ & $0.90$  \\
    & CAML & $\mathbf{0.66}$ & $\mathbf{0.60}$ & $\mathbf{0.54}$ && $\mathbf{0.44}$ & $\mathbf{0.61}$ & $\mathbf{0.54}$ && $\mathbf{0.55}$ & $\mathbf{0.63}$ & $\mathbf{0.76}$ \\
		   	\hline
		   	\multirow{11}{*}{PSNR$\uparrow$}
            & PICNet$^\sharp$ & $23.59$ & $21.70$ & $16.29$ && $26.84$ & $23.29$ & $18.80$ && $25.34$ & $20.84$ & $16.05$\\
		   	& CTSDG$^\sharp$ & $27.58$ & $24.25$ & $20.05$ && $28.86$ & $24.35$ & $20.36$ && $25.93$ & $22.00$ & $18.36$ \\ 
      	& LaMa$^\sharp$  & $26.30$ & $23.41$ & $19.91$ && $30.61$ & $26.07$ & $21.89$ && $25.90$ & $22.11$ & $18.64$ \\ 
        & MISF$^\sharp$ & $26.84$ & $24.49$ & $19.47$&& $30.06$ & $25.07$ & $19.91$ && $27.06$ & $22.54$ & $18.42$ \\ 
        & CoordFill$^\sharp$ & $27.01$ & $24.29$ & $20.88$ && $26.92$ & $24.66$ & $21.35$ && $26.69$ & $22.84$ & $19.19$ \\ 
    &Repaint$^\sharp$ & $26.51$ & $24.83$ & $20.57$ && $30.23$ & $25.14$ & $22.28$ && $25.45$ & $24.41$ & $21.74$\\
    & Uni-paint$^\sharp$ & $24.53$ & $23.19$ & $20.08$ & & $27.68$ & $23.72$ & $20.65$ && $22.55$ & $21.14$ & $20.72$\\
    & VCN & $24.26$ & $22.27$ & $18.93$ && $28.28$ & $24.81$ & $20.94$ && $25.49$ & $22.25$ & $18.84$ \\
    & TransCNN-HAE & $27.42$ & $25.13$ & $21.46$ && $30.03$ & $26.08$ & $22.06$ && $26.85$ & $23.78$ & $20.65$ \\
    & OmniWavNet & $24.77$ & $22.49$ & $19.72$ && $27.35$ & $24.65$ & $20.04$&& - & - & - \\
    & CAML & $\mathbf{27.63}$& $\mathbf{25.18}$ & $\mathbf{21.97}$ && $\mathbf{30.90}$ & $\mathbf{26.27}$ & $\mathbf{22.90}$&
    & $\mathbf{27.17}$ & $\mathbf{25.41}$ & $\mathbf{23.58}$ \\
		   	\hline
		   	\multirow{11}{*}{SSIM$\uparrow$}
            & PICNet$^\sharp$ & $0.864$ & $0.834$ & $0.575$ && $0.942$ & $0.885$ & $0.741$ && $0.916$ & $0.790$ & $0.527$ \\
		   	& CTSDG$^\sharp$ & $0.952$ & $0.903$ & $0.774$ && $0.962$ & $0.901$ & $0.779$ && $0.925$ & $0.828$ & $0.640$ \\ 
      	& LaMa$^\sharp$ & $0.954$ & $0.902$ & $0.790$ && $0.975$ & $0.936$ & $0.852$ && $0.929$ & $0.840$ & $0.668$ \\ 
        & MISF$^\sharp$ & $0.950$ & $0.906$ & $0.750$ && $0.971$ & $0.916$ & $0.776$ && $0.942$ & $0.847$ & $0.652$ \\ 
        & CoordFill$^\sharp$ & $0.911$ & $0.891$ & $0.798$ && $0.941$ & $0.906$ & $0.816$ && $0.933$ & $0.848$ & $0.678$ \\ 
        &Repaint$^\sharp$ & $0.928$ & $0.903$ & $0.809$ & & $0.972$ & $0.931$ & $0.855$ & & $0.919$ & $0.853$& $0.788$\\
    & Uni-paint$^\sharp$ & $0.821$ & $0.767$ & $0.673$ && $0.834$ & $0.711$ & $0.726$ && $0.830$ & $0.818$ & $0.691$ \\  
    & VCN & $0.912$ & $0.865$ & $0.745$ && $0.957$ & $0.910$ & $0.808$ && $0.914$ & $0.829$ & $0.666$ \\
    & TransCNN-HAE & $0.951$ & $0.919$ & $0.829$ && $0.971$ & $0.932$ & $0.850$ && $0.935$ & $0.872$ & $0.753$\\
    & OmniWavNet & $0.861$ & $0.869$ & $0.721$ && $0.952$ & $0.910$ & $0.804$&& - & - & - \\
    & CAML & $\mathbf{0.956}$& $\mathbf{0.925}$ & $\mathbf{0.851}$ && $\mathbf{0.978}$ & $\mathbf{0.937}$ & $\mathbf{0.870}$ && $\mathbf{0.957}$ & $\mathbf{0.918}$ & $\mathbf{0.820}$ \\
		   	\hline
		   	\multirow{11}{*}{FID$\downarrow$}
            & PICNet$^\sharp$ & $14.14$ & $22.72$ & $65.52$ && $5.80$ & $8.85$ & $19.04$ && $9.64$ & $29.27$ & $97.62$ \\
		   	& CTSDG$^\sharp$ & $7.18$ & $11.41$ & $32.06$ && $4.44$ & $9.41$ & $21.64$ && $8.65$ & $21.46$ & $62.60$ \\ 
      	& LaMa$^\sharp$  & $4.81$ & $8.27$ & $17.67$ && $2.36$ & $4.98$ & $9.39$ && $4.82$ & $13.13$ & $34.15$ \\ 
        & MISF$^\sharp$ & $6.23$ & $7.70$ & $20.03$ && $2.75$ & $6.22$ & $16.06$ && $4.45$ & $12.94$ & $41.65$ \\ 
        & CoordFill$^\sharp$ & $5.47$ & $7.97$ & $21.05$ && $16.36$ & $16.59$ & $21.79$ && $6.41$ & $15.37$ & $38.92$  \\ 
    &Repaint$^\sharp$ & $9.07$ & $7.30$ & $12.99$ && $4.21$ & $7.42$ & $10.65$ && $4.56$ & $16.14$ & $31.06$ \\
    & Uni-paint$^\sharp$ & $19.16$ & $18.12$ & $21.30$ && $10.81$ & $13.54$ & $18.73$ && $12.39$ & $18.19$ & $40.58$  \\
    & VCN & $7.95$ & $10.71$ & $21.32$ && $4.33$ & $9.00$  & $20.95$ && $8.29$ & $22.39$ & $62.11$ \\
    & TransCNN-HAE & $4.70$ & $7.15$ & $14.55$ && $2.92$  & $6.08$ & $13.23$ && $6.32$ & $13.46$ & $32.64$ \\
    & OmniWavNet & $9.43$ & $10.61$ & $23.95$ && $5.86$ & $8.12$ & $21.16$&& - & - & - \\
    & CAML & $\mathbf{4.68}$ & $\mathbf{6.69}$ & $\mathbf{11.36}$ && $\mathbf{2.26}$ & $\mathbf{4.67}$ &  $\mathbf{9.05}$ && $\mathbf{4.07}$ & $\mathbf{12.24}$ & $\mathbf{30.29}$ \\
		   	\hline
		   	\multirow{11}{*}{LPIPS$\downarrow$}
            & PICNet$^\sharp$ & $0.136$ & $0.151$ & $0.345$ && $0.078$ & $0.122$ & $0.231$ && $0.165$ & $0.256$ & $0.434$  \\
		   	& CTSDG$^\sharp$ & $0.068$ & $0.103$ & $0.220$ && $0.055$ & $0.118$ & $0.230$ && $0.153$ & $0.233$ & $0.376$\\ 
      	& LaMa$^\sharp$ & $0.045$ & $0.090$ & $0.179$ && $0.028$ & $\mathbf{0.062}$ & $0.127$ && $0.115$ & $0.178$ & $0.296$ \\ 
        & MISF$^\sharp$ & $0.069$ & $0.086$ & $0.210$ && $0.034$ & $0.079$ & $0.191$ && $0.108$ & $0.153$ & $0.305$ \\ 
        & CoordFill$^\sharp$ & $0.061$ & $0.085$ & $0.174$ && $0.151$ & $0.160$ & $0.201$ && $0.138$ & $0.198$ & $0.312$ \\ 
         &Repaint$^\sharp$ & $0.213$ & $0.123$ & $0.235$ & & $0.102$ & $0.215$ & $0.127$ && $0.143$ & $0.201$ & $0.256$  \\
    & Uni-paint$^\sharp$ & $0.249$ & $0.201$ & $0.227$ && $0.158$ & $0.174$ & $0.182$ && $0.152$ & $0.195$ & $0.212$\\
    & VCN & $0.097$ & $0.131$ & $0.219$ && $0.052$ & $0.099$ & $0.191$ && $0.131$ & $0.209$ & $0.351$ \\
    & TransCNN-HAE & $0.047$ & $0.080$ & $0.156$ && $0.033$ & $0.071$ & $0.143$ && $0.132$ & $0.185$ & $0.281$ \\
    & OmniWavNet & $0.201$ & $0.147$ & $0.237$ && $0.054$ & $0.095$ & $0.234$ && - & - & - \\
    & CAML & $\mathbf{0.042}$& $\mathbf{0.076}$ & $\mathbf{0.141}$ && $\mathbf{0.026}$ & $\mathbf{0.062}$ & $\mathbf{0.113}$ && $\mathbf{0.083}$ & $\mathbf{0.116}$ & $\mathbf{0.202}$ \\
		   	\hline
\multicolumn{13}{l}{$\uparrow$ indicates the higher the better, and $\downarrow$ indicates the lower the better. \textbf{Bold} means the best results.}\\

		\end{tabular}
	  \caption{Quantitative comparison results on FFHQ, CelebA-HQ, and ImageNet datasets. $\sharp$ means the non-blind image inpainting methods with ground truth masks, and otherwise means the blind image inpainting methods.}
\label{tab:blind1}
\end{table*}

\textbf{Metrics.}
Following~\citep{wang2020vcnet,zhao2022transcnn}, we use Binary Cross-Entropy (BCE) error to evaluate the performance of mask estimation. BCE measures the accuracy of estimated masks. We use image fidelity evaluation metrics Peak Signal-to-Noise Ratio (PSNR) and Structural Similarity (SSIM) as well as image perceptual quality evaluation metrics Frechet Inception Distance (FID) and Learned Perceptual Image Patch Similarity (LPIPS) to evaluate the performance of image inpainting. PSNR and SSIM measure the pixel-wise quality of the recovered image. FID measures the perceptual quality of inpainting results. LPIPS measures the perceptual similarity between real images and inpainting images for evaluating reasonability.


\begin{table*}[t!]
 \centering
\scriptsize
\setlength{\tabcolsep}{3mm}
	\begin{tabular}{c c c c c c c c c}
			\hline
			\multicolumn{1}{c}{\multirow{2}{*}{Metric}} &
			\multicolumn{1}{c}{\multirow{2}{*}{Method}} &
			\multicolumn{3}{c}{\multirow{1}{*}{Paris StreetView}} &&
			\multicolumn{3}{c}{\multirow{1}{*}{Places2}}
			\\
      \cline{3-5}  \cline{7-9} 
   & & $0$-$20\%$ & $20$-$40\%$ & $40$-$60\%$ && $0$-$20\%$ & $20$-$40\%$ & $40$-$60\%$ 
   \\
		   	\hline
		   	\multirow{2}{*}{BCE$\downarrow$}
    & VCN & $0.70$ & $0.83$ & $0.92$ && $0.66$ & $0.81$ & $0.91$ \\
    & CAML & $\mathbf{0.66}$ & $\mathbf{0.62}$ & $\mathbf{0.55}$ && $\mathbf{0.41}$ & $\mathbf{0.60}$ & $\mathbf{0.53}$\\
		   	\hline
		   	\multirow{11}{*}{PSNR$\uparrow$}
            & PICNet$^\sharp$ & $26.73$ & $23.26$ & $19.03$ && $27.25$ & $22.61$ & $17.98$ \\
		   	& CTSDG$^\sharp$ & $\mathbf{30.02}$ & $25.87$ & $21.76$ && $29.60$ & $25.12$ & $21.28$\\ 
      	& LaMa$^\sharp$ & $27.70$ & $24.84$ & $21.32$ && $29.13$ & $25.17$ & $21.64$\\ 
        & MISF$^\sharp$ & $28.23$ & $25.00$ & $20.14$ && $30.22$ & $25.67$ & $21.60$\\ 
        & CoordFill$^\sharp$ & $27.88$ & $24.70$ & $20.21$ && $25.86$ & $24.00$ & $21.39$\\ 
          &Repaint$^\sharp$ & $\mathbf{29.10}$ & $25.02$ & $22.76$ && $28.31$ & $24.77$ & $22.65$\\
    & Uni-paint$^\sharp$ & $25.68$ & $24.51$ & $21.49$ && $27.74$ & $22.70$ & $21.23$ \\
    & VCN & $27.00$ & $24.34$ & $21.02$ && $28.80$ & $25.30$ & $21.74$\\
    & TransCNN-HAE & $29.89$ & $25.44$ & $22.48$ && $\mathbf{30.48}$ & $26.05$ & $22.93$\\
    & OmniWavNet & $26.71$ & $23.29$ & $19.25$ && $27.77$ & $24.55$ & $21.83$\\
    & CAML & $28.89$ & $\mathbf{27.63}$ & $\mathbf{23.59}$ && $29.31$&  $\mathbf{26.44}$ & $\mathbf{23.54}$\\
		   	\hline
		   	\multirow{11}{*}{SSIM$\uparrow$}
            & PICNet$^\sharp$ & $0.909$ & $0.815$ & $0.591$ && $0.922$ & $0.806$ & $0.570$\\
		   	& CTSDG$^\sharp$ & $\mathbf{0.955}$ & $0.882$ & $0.727$  && $0.949$ & $0.871$ & $0.723$\\ 
      	& LaMa$^\sharp$ & $0.951$ & $0.881$ & $0.733$ && $0.949$ & $0.878$ & $0.743$\\ 
        & MISF$^\sharp$ & $0.938$ & $0.864$ & $0.664$ && $0.952$ & $0.880$ & $0.734$\\ 
        & CoordFill$^\sharp$ & $0.932$ & $0.878$ & $0.749$ && $0.882$ & $0.825$ & $0.707$\\ 
              &Repaint$^\sharp$ & $0.942$ & $0.892$ & $0.806$ && $0.935$ & $0.850$ & $0.802$ \\
    & Uni-paint$^\sharp$ & $0.810$ & $0.842$ & $0.736$ && $0.879$ & $0.838$ & $0.684$\\
    & VCN & $0.905$ & $0.832$ & $0.685$ && $0.938$ & $0.869$ & $0.735$\\
    & TransCNN-HAE & $0.954$ & $0.893$ & $0.767$ && $\mathbf{0.955}$ & $0.887$ & $0.783$\\
    & OmniWavNet & $0.919$ & $0.813$ & $0.599$ && $0.909$ & $0.811$ & $0.727$\\
    & CAML  & $0.945 $ & $\mathbf{0.922}$ & $\mathbf{0.826}$ && $0.954$ & $\mathbf{0.899}$ & $\mathbf{0.813}$\\
		   	\hline
		   	\multirow{11}{*}{FID$\downarrow$}
            & PICNet$^\sharp$ & $37.03$ & $62.91$ & $112.07$ && $16.10$ & $33.38$ & $71.93$\\
		   	& CTSDG$^\sharp$ & $22.08$ & $49.42$ & $98.77$ && $12.45$ & $26.00$ & $49.05$\\ 
      	& LaMa$^\sharp$  & $23.42$ & $49.42$ & $84.30$ && $10.05$ & $18.42$ & $32.17$\\ 
        & MISF$^\sharp$ & $33.11$ & $50.08$ & $110.35$ && $9.35$ & $18.72$ & $33.00$\\ 
        & CoordFill$^\sharp$ & $38.35$ & $41.39$ & $93.62$ && $37.80$ & $42.38$ & $54.55$\\ 
                &Repaint$^\sharp$ & $21.24$ & $35.83$ & $72.26$ && $10.32$ & $20.48$ & $32.39$\\
    & Uni-paint$^\sharp$ & $23.77$ & $44.12$ & $80.75$ && $16.72$ & $23.63$ & $50.75$\\
    & VCN & $47.52$ & $72.13$ & $111.39$ && $13.94$ & $27.37$ & $54.15$\\
    & TransCNN-HAE & $21.44$ & $42.70$ & $81.58$ && $10.32$ & $20.82$ & $40.03$\\
    & OmniWavNet & $38.44$ & $79.47$ & $138.07$ && $15.33$ & $27.99$ & $56.72$\\
    & CAML & $\mathbf{14.95}$ & $\mathbf{31.16}$ & $\mathbf{68.03}$ && $\mathbf{9.29}$ & $\mathbf{18.06}$ & $\mathbf{30.20}$\\
		   	\hline
		   	\multirow{12}{*}{LPIPS$\downarrow$}
            & PICNet$^\sharp$ & $0.109$ & $0.180$ & $0.349$ && $0.082$ & $0.185$ & $0.384$ \\
		   	& CTSDG$^\sharp$ & $0.052$ & $0.119$ & $0.246$ && $0.062$ & $0.139$ & $0.277$\\ 
      	& LaMa$^\sharp$ & $0.053$ & $0.110$ & $0.221$ && $0.047$ & $0.097$ & $0.193$\\ 
        & MISF$^\sharp$ & $0.090$ & $0.127$ & $0.282$ && $0.050$ & $0.100$ & $0.203$\\ 
        & CoordFill$^\sharp$ & $0.118$ & $0.201$ & $0.243$ && $0.211$ & $0.226$ & $0.273$\\
                &Repaint$^\sharp$ & $0.128$ & $0.202$ & $0.245$ && $0.151$ & $0.206$ & $0.241$\\
    & Uni-paint$^\sharp$ & $0.152$ & $0.271$ & $0.252$ && $0.238$ & $0.224$ & $0.286$\\
    & VCN & $0.113$ & $0.180$ & $0.313$ && $0.071$ & $0.146$ & $0.292$\\
    & TransCNN-HAE & $0.049$ & $0.102$ & $0.203$ && $0.050$ & $0.105$ & $0.203$\\
    & OmniWavNet & $0.095$ & $0.205$ & $0.408$ && $0.169$ & $0.124$ & $0.261$\\
    & CAML & $\mathbf{0.031}$ & $\mathbf{0.076}$  & $\mathbf{0.167}$ && $\mathbf{0.043}$ & $\mathbf{0.093}$ & $\mathbf{0.180}$\\
		   	\hline
\multicolumn{9}{l}{$\uparrow$ indicates the higher the better, and $\downarrow$ indicates the lower the better. \textbf{Bold} means the best results.}\\
		\end{tabular}
	  \caption{Quantitative comparison results on Paris StreetView and Places2 datasets.}
	  \label{tab:blindimageinpainting2}
\end{table*}

\textbf{State-of-the-arts.}
We compare our method with state-of-the-art blind inpainting methods VCN~\citep{wang2020vcnet}, TransCNN-HAE~\citep{zhao2022transcnn}, and OmniWavNet~\citep{phutke2023blind} as well as state-of-the-art non-blind image inpainting methods PICNet~\citep{zheng2019pluralistic}, CTSDG~\citep{guo2021image}, LaMa~\citep{suvorov2022resolution}, MISF~\citep{li2022misf}, CoordFill~\citep{liu2023coordfill}, Repaint~\cite{lugmayr2022repaint}, and Uni-paint~\cite{yang2023uni}. 
All employed state-of-the-art blind and non-blind methods have publicly accessible training codes or pre-trained models.
For the purpose of fairness, we try to use officially released pre-trained models directly with the same experimental settings as much as possible. For some methods and datasets with unreleased pre-trained models, we use officially released training codes to retrain models until convergence following the same experimental settings proposed in each method. Since OmniWavNet only provides pre-trained models on FFHQ, CelebA-HQ, Paris StreetView, and Places2 datasets without releasing training codes, our test is limited to these datasets for OmniWavNet.
Notably, for non-blind image inpainting methods, since it is not suitable for blind image inpainting, we equip the ground truth mask $\mathbf{M}$ as input for them during training and testing.

\begin{figure}[!t]
\centering
\includegraphics[width=\linewidth]{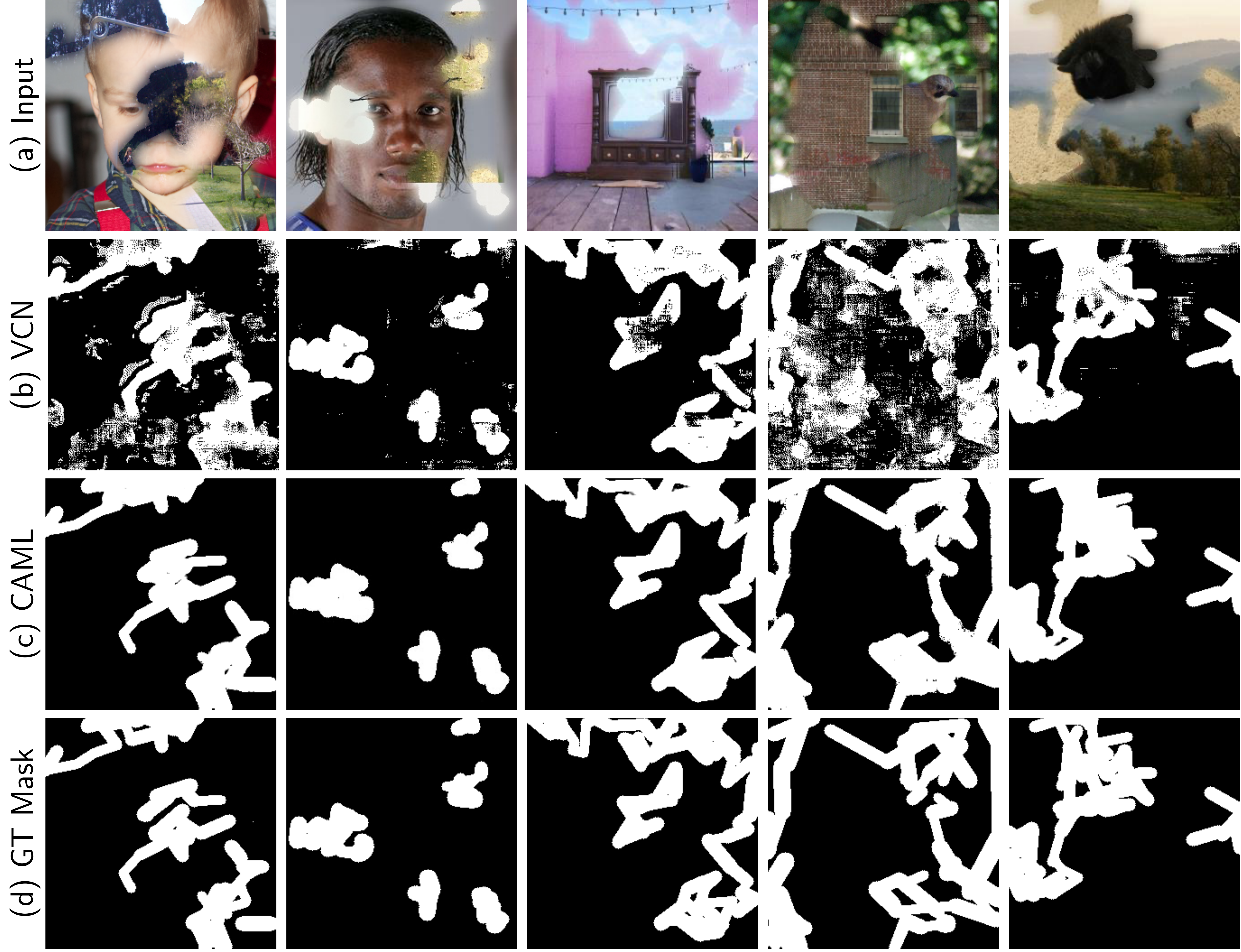}
\caption{Qualitative comparison results of mask estimation for blind image inpainting: (a) Input contaminated images, (b) VCN, (c) our CAML, and (d) Ground Truth Mask (GT Mask). From left to right (one example for each dataset): FFHQ, CelebA-HQ, ImageNet, Paris StreetView, and Places2.}
\label{fig:QualitativeComparisonMask}
\end{figure}

\begin{figure*}[!t]
\centering
\includegraphics[width=\linewidth]{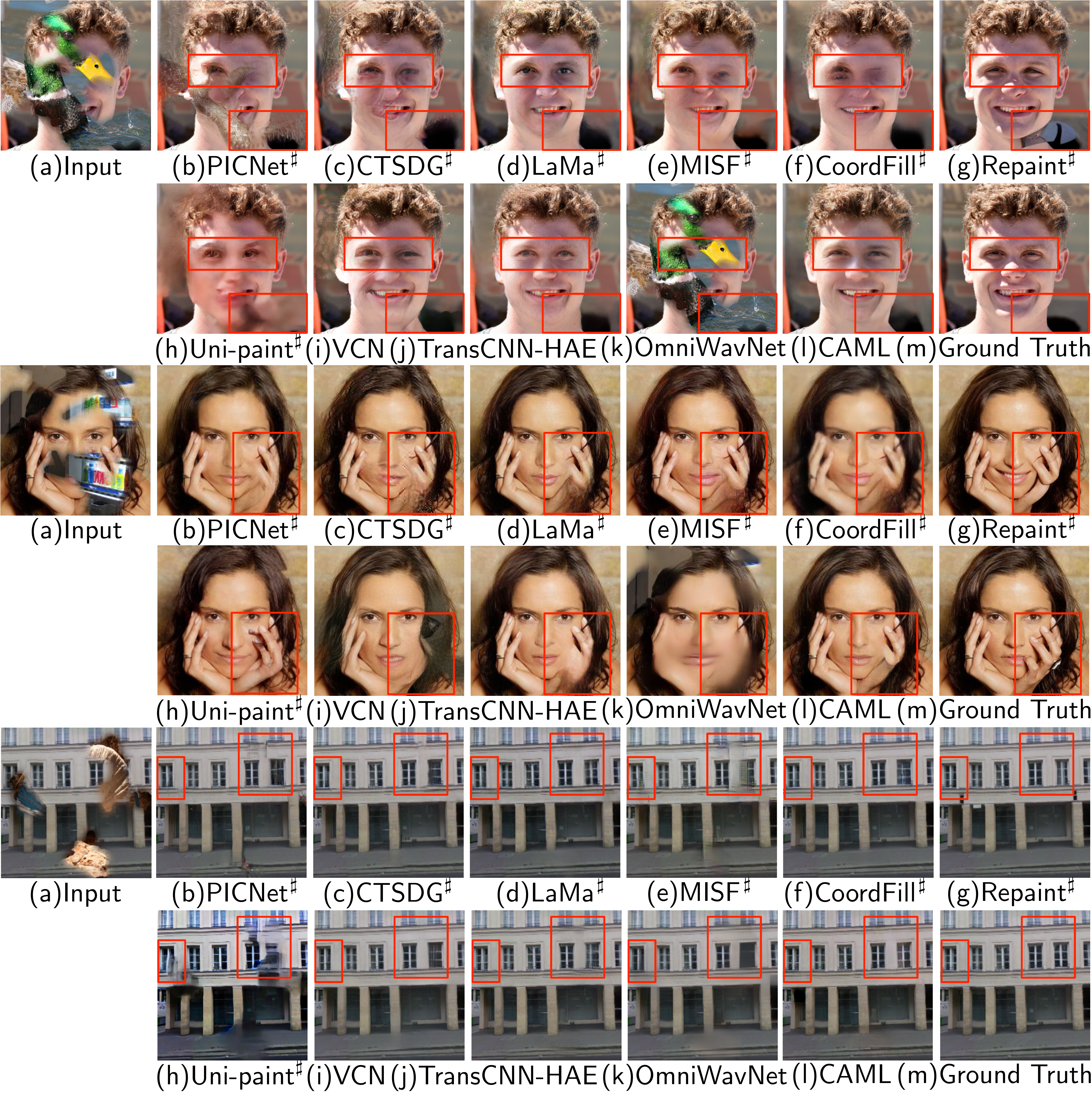}
\caption{Qualitative comparison results of image inpainting for blind image inpainting. From top to bottom (one example for each dataset): FFHQ, CelebA-HQ, and Paris StreetView.}
\label{fig:QualitativeComparisonInpainting1}
\end{figure*}

\begin{figure*}[!t]
\centering
\includegraphics[width=\linewidth]{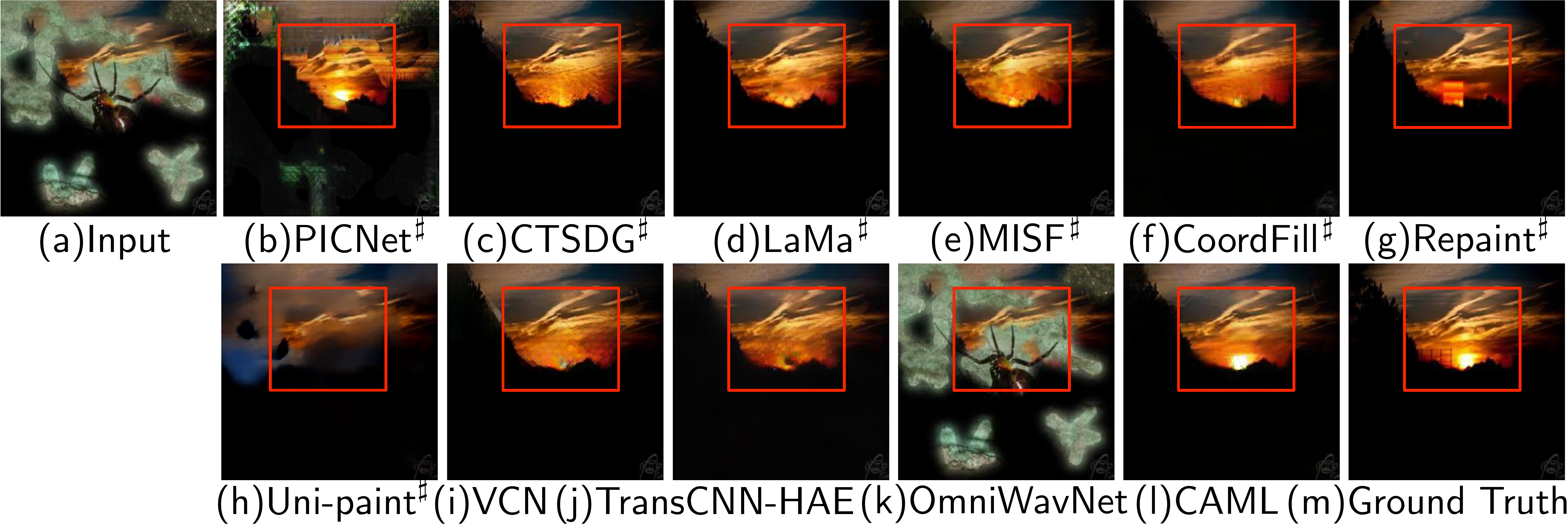}
\caption{Qualitative comparison results of image inpainting for blind image inpainting on Places2.}
\label{fig:QualitativeComparisonInpainting2}
\end{figure*}

\begin{figure}[!t]
\centering
\includegraphics[width=\linewidth]{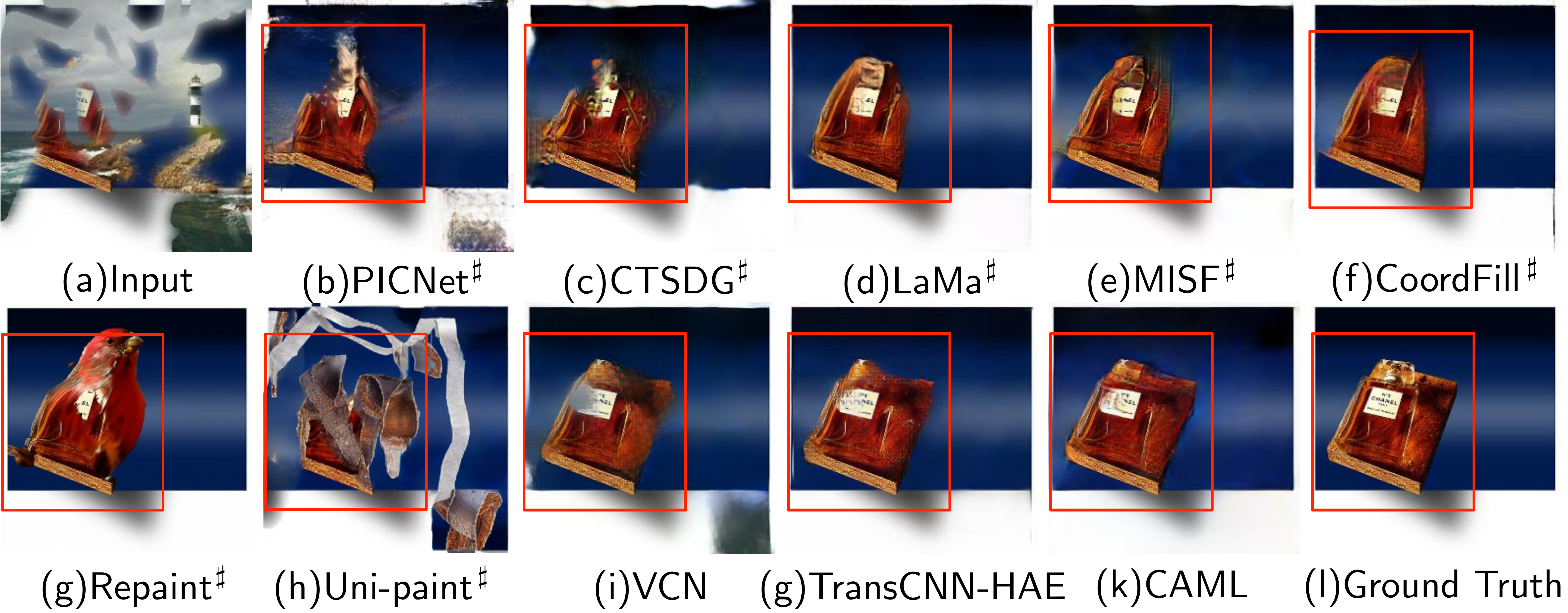}
\caption{Qualitative comparison results of image inpainting for blind image inpainting on ImageNet.}
\label{fig:QualitativeComparisonInpainting3}
\end{figure}

\subsection{Performance Evaluation}
\textbf{Quantitative Comparison.} 
We provide quantitative comparison results on the five public datasets with mask rations $0$-$20\%$, $20$-$40\%$, and $40$-$60\%$ to compare our CAML with state-of-the-art blind (VCN, TransCNN-HAE, and OmniWavNet) and non-blind image inpainting (PICNet, CTSDG, LaMa, MISF,  CoordFill, Repaint, and Uni-paint) methods, as shown in Table~\ref{tab:blind1} and Table~\ref{tab:blindimageinpainting2}. The results demonstrate that our CAML achieves superior performance in mask estimation and accurately predicts masks of different mask rations compared to the two-stage blind image inpainting method VCN. Meanwhile, our CAML achieves consistent and significant performance gains in image inpainting, especially when compared with the non-blind methods with the ground truth mask as input. It is worth noting that under small mask rations ($0$-$20\%$), the performance of our method in terms of fidelity metrics (PSNR and SSIM) is not as prominent as its performance in terms of perceptual metrics (FID and LPIPS). However, preliminary works~\citep{ledig2017photo, sajjadi2017enhancenet} have indicated that the perceptual metrics both are accurate in reflecting the realism of images and are consistent with the human evaluation more than the fidelity metrics. Moreover, as the mask ratio increases, our CAML exhibits a more pronounced performance advantage, thereby demonstrating the robustness under different mask ratios.

\textbf{Qualitative Comparison.} 
We next show qualitative comparison results of mask estimation by comparing our CAML with VCN across all datasets in Figure~\ref{fig:QualitativeComparisonMask}, indicating that our CAML can estimate more accurate masks.
We also show qualitative comparison results of image inpainting for different methods across all datasets in Figure~\ref{fig:QualitativeComparisonInpainting1}, Figure~\ref{fig:QualitativeComparisonInpainting2} and Figure~\ref{fig:QualitativeComparisonInpainting3}. It can be seen that, our CAML can generate visually plausible images.   Particularly in cases of large-scale contamination, both non-blind methods and some blind methods may result in distorted content (\eg, eyes, mouth, and hands in face samples), diffusion model-based non-blind methods (Repaint and Uni-paint) are relatively unstable, and OmniWavNet even fails to remove the contamination. Our CAML yields more reasonable and realistic content.

\begin{table*}[!t]
\footnotesize
    \setlength{\tabcolsep}{1mm}
    \centering
		\begin{tabular}{c c c c c c c c}
			\hline
 \multicolumn{2}{c}{Method} && FFHQ & CelebA-HQ & ImageNet & Paris StreetView & Places2 \\    
        \cline{1-2}       \cline{4-8}
 \multirow{10}{*}{CAML} 
 & $>$PICNet$^\sharp$ && $89.75\%$ & $81.25\%$ & $78.50\%$ & $84.00\%$ & $85.25\%$\\
          &  $>$CTSDG$^\sharp$ && $87.25\%$ & $72.75\%$ & $87.25\%$ & $89.75\%$ & $86.50\%$\\
         &  $>$LaMa$^\sharp$ &&  $76.25\%$ & $68.50\%$ & $80.50\%$ & $82.75\%$ & $81.75\%$ \\
         &  $>$MISF$^\sharp$ && $89.50\%$ & $84.00\%$ & $88.00\%$ & $85.25\%$ & $91.75\%$ \\
         & $>$CoordFill$^\sharp$ && $79.00\%$ & $71.75\%$ & $86.75\%$ & $87.50\%$ & $72.75\%$ \\
         &  $>$Repaint$^\sharp$ && $90.00\%$ & $85.50\%$ & $87.25\%$ & $85.50\%$ & $90.50\%$ \\
         & $>$Uni-paint$^\sharp$ && $80.25\%$ & $75.00\%$ & $65.25\%$ & $85.50\%$ & $80.00\%$ \\
       &  $>$VCN && $90.75\%$ & $84.00\%$ & $94.75\%$ & $87.75\%$ & $86.25\%$ \\
       &   $>$TransCNN-HAE  && $77.75\%$ & $73.75\%$ & $85.25\%$ & $80.75\%$ & $88.00\%$\\
        &    $>$OmniWavNet &&  $91.25\%$ & $90.00\%$ & $97.50\%$ & $94.00\%$ & $95.25\%$ \\
        
\hline 
		\end{tabular}
	   \caption{User study results. Each entry shows the percentage of cases where the proposed method is judged as more realistic and reasonable than all existing competitors. $\sharp$ means the non-blind image inpainting methods with ground truth masks, and otherwise means the blind image inpainting methods.}
	  \label{tab:user_study}
\end{table*}

\textbf{User Study.}
We conduct the human visual evaluation on five public datasets for a more credible conclusion, due to a gap between the evaluation metrics and human perception. We employ the pairwise A/B tests following~\citep{wang2020vcnet, zhao2022transcnn} that each volunteer is only required to select the more realistic and reasonable image from pairwise comparisons within unlimited time. Specifically, we invite $50$ volunteers to evaluate $16$ questionnaires. Every questionnaire involves $40$ pairwise comparisons, and every comparison shows results predicted from two different methods based on the same input with the same random masks, randomized in the left-right order. For non-blind image inpainting methods, we equip the ground truth mask as the input. Table~\ref{tab:user_study} shows that our CAML achieves the best performance across all datasets.

\begin{table*}[!t]
\scriptsize
    \setlength{\tabcolsep}{0.2mm}
    \centering
		\begin{tabular}{c c c c c c c c c c c c }
			\hline
			\multicolumn{1}{c}{\multirow{2}{*}{Dataset}} &
   			\multicolumn{1}{c}{\multirow{2}{*}{Contaminated Pattern}} &
			\multicolumn{5}{c}{\multirow{1}{*}{VCN}} &&
   			\multicolumn{4}{c}{\multirow{1}{*}{TransCNN-HAE}}  \\
      \cline{3-7} \cline{9-12}
           & &  BCE$\downarrow$ $\downarrow$ & PSNR$\uparrow$ & SSIM$\uparrow$ & FID$\downarrow$ & LPIPS$\downarrow$
          && PSNR$\uparrow$ & SSIM$\uparrow$ & FID$\downarrow$  & LPIPS$\downarrow$ \\    
           \hline
            \multirow{3}{*}{FFHQ} & Graffiti & $0.756$ & $21.04$ & $0.797$ & $19.05$ & $0.232$ && $22.77$ & $0.813$ & $14.90$ & $0.182$\\
            &  Text & $0.751$  & $21.52$ & $0.848$ & $21.41$ & $0.157$ && $23.98$ & $0.905$ & $21.64$ & $0.144$ \\
            & Image Occlusion & $0.755$ & $20.32$ & $0.815$ & $61.84$ & $0.421$ && $23.88$ & $0.895$ & $56.06$ & $0.257$ \\
         \hline
           \multirow{3}{*}{CelebA-HQ} & Graffiti & $0.715$ &  $19.08$ & $0.746$ & $21.11$ & $0.298$ && $21.19$ & $0.814$ & $13.81$ & $0.246$\\
           &  Text & $0.708$  &  $23.10$ & $0.887$ & $10.26$ & $0.174$ && $24.58$ & $0.911$ & $10.81$ & $0.158$\\
           & Image Occlusion & $0.716$ & $25.08$ & $0.918$ & $9.22$ & $0.121$ && $26.16$ & $0.926$ & $7.14$ & $0.105$ \\
         \hline
           \multirow{3}{*}{ImageNet} & Graffiti & $0.714$ & $20.13$ & $0.789$ & $24.32$ & $0.106$ &&$22.09$ & $0.799$ & $21.05$ & $0.089$ \\
         &  Text & $0.700$ & $25.23$ & $0.917$ & $13.31$ & $0.188$ && $26.01$ & $0.927$ & $8.65$ & $0.131$ \\
            & Image Occlusion & $0.698$ & $19.85$ & $0.834$ & $29.45$ & $0.192$ && $22.53$ & $0.857$ & $18.45$  & $0.110$\\
       \hline
           \multirow{3}{*}{Paris StreetView} & Graffiti & $0.753$  & $23.11$ & $0.793$ & $70.38$ & $0.202$ && $27.37$ & $0.905$ & $53.58$ & $0.139$ \\
           &  Text & $0.731$ & $27.00$ & $0.916$ & $40.75$ & $0.262$ && $33.46$  & $0.971$ & $24.72$ & $0.117$\\
           & Image Occlusion & $0.742$ & $25.86$ & $0.860$ & $49.44$ & $0.274$ && $26.51$ & $0.864$ & $42.49$ & $0.213$ \\    
    \hline
           \multirow{3}{*}{Places2} & Graffiti & $0.720$ & $18.37$ & $0.648$ & $64.28$ & $0.387$ && $20.63$ & $0.712$ & $58.08$ & $0.276$\\
               &  Text & $0.723$ &$26.56$ & $0.908$ & $21.00$  & $0.144$ &&$28.52$ & $0.939$ & $17.56$ & $0.118$ \\
           & Image Occlusion & $0.728$ &$21.69$ & $0.797$ & $24.56$ & $0.122$ &&$23.87$ & $0.859$ & $18.73$ & $0.115$\\    
         \hline
         
    \end{tabular} 
          \begin{tabular}{cc c c c c c c c c c c }
			\multicolumn{1}{c}{\multirow{2}{*}{Dataset}} &
   			\multicolumn{1}{c}{\multirow{2}{*}{Contaminated Pattern}} &
			\multicolumn{4}{c}{\multirow{1}{*}{OmniWavNet}} &&
   			\multicolumn{5}{c}{\multirow{1}{*}{CAML}}  \\
             \cline{3-6} \cline{8-12}
           &  & PSNR$\uparrow$ & SSIM$\uparrow$ & FID$\downarrow$ & LPIPS$\downarrow$
           && BCE$\downarrow$ & PSNR$\uparrow$ & SSIM$\uparrow$ & FID$\downarrow$  & LPIPS$\downarrow$ \\   
 \hline
            \multirow{3}{*}{FFHQ} & Graffiti & $20.87$ & $0.791$ & $18.79$ & $0.318$ &&  $\mathbf{0.694}$ &  $\mathbf{23.33}$ & $\mathbf{0.882}$ & $\mathbf{12.22}$ & $\mathbf{0.121}$\\
            &  Text   & $21.04$  & $0.793$ & $25.58$ & $0.202$ && $\mathbf{0.693}$  & $\mathbf{24.40}$ & $\mathbf{0.928}$ & $\mathbf{18.13}$ & $\mathbf{0.137}$ \\
            & Image Occlusion & $20.85$ & $0.807$ & $68.34$ & $0.393$ && $\mathbf{0.693}$ & $\mathbf{25.19}$ & $\mathbf{0.930}$ & $\mathbf{44.68}$ & $\mathbf{0.219}$\\
         \hline
           \multirow{3}{*}{CelebA-HQ} & Graffiti  & $20.71$ &  $0.798$& $11.35$ & $0.282$ && $\mathbf{0.681}$ & $\mathbf{21.52}$ & $\mathbf{0.824}$ & $\mathbf{10.89}$ & $\mathbf{0.221}$\\
           &  Text  & $23.86$ & $0.860$ & $12.11$ & $0.171$ && $\mathbf{0.691}$ & $\mathbf{25.65}$ & $\mathbf{0.927}$ & $\mathbf{8.48}$ & $\mathbf{0.125}$ \\
           & Image Occlusion & $25.44$& $0.893$ & $10.12$ &  $0.164$ && $\mathbf{0.678}$ & $\mathbf{26.50}$ & $\mathbf{0.945}$ & $\mathbf{5.53}$ & $\mathbf{0.074}$ \\
         \hline
           \multirow{3}{*}{ImageNet} & Graffiti & - &  -& -& - && $\mathbf{0.661}$  & $\mathbf{24.72}$ & $\mathbf{0.820}$ & $\mathbf{19.21}$ & $\mathbf{0.062}$  \\
         &  Text  & - &  -& -& - && $\mathbf{0.663}$ & $\mathbf{26.49}$ & $\mathbf{0.933}$ & $\mathbf{7.44}$ & $\mathbf{0.126}$\\
            & Image Occlusion  & - &  -& -& - && $\mathbf{0.663}$ & $\mathbf{23.16}$ & $\mathbf{0.885}$ & $\mathbf{12.59}$ & $\mathbf{0.084}$ \\
       \hline
           \multirow{3}{*}{Paris StreetView} & Graffiti  & $27.56$ &  $0.883$& $55.22$ & $0.251$ && $\mathbf{0.674}$ & $\mathbf{28.33}$ & $\mathbf{0.924}$ & $\mathbf{49.71}$ & $\mathbf{0.108}$\\
           &  Text & $30.52$ & $0.958$ & $27.36$ & $0.224$ && $\mathbf{0.690}$ & $\mathbf{34.45}$ & $\mathbf{0.974}$ & $\mathbf{23.87}$ & $\mathbf{0.103}$ \\
           & Image Occlusion  & $25.17$ & $0.853$ & $49.27$  & $0.216$ && $\mathbf{0.687}$  & $\mathbf{28.01}$ & $\mathbf{0.892}$ & $\mathbf{39.65}$ & $\mathbf{0.165}$ \\    
    \hline
           \multirow{3}{*}{Places2} & Graffiti & $20.80$ & $0.691$& $61.32$ & $0.308$ &&  $\mathbf{0.672}$ & $\mathbf{21.54}$ & $\mathbf{0.744}$ & $\mathbf{49.26}$
           & $\mathbf{0.228}$\\
               &  Text  & $26.27$ & $0.889$ & $22.14$ & $0.193$  && $\mathbf{0.691}$  & $\mathbf{29.07}$ & $\mathbf{0.950}$ & $\mathbf{15.52}$ & $\mathbf{0.107}$ \\
           & Image Occlusion   & $23.49$ & $0.811$ & $20.32$ & $0.136$ && $\mathbf{0.679}$ & $\mathbf{25.45}$ & $\mathbf{0.908}$ & $\mathbf{16.18}$ & $\mathbf{0.096}$\\    
           \hline
		\end{tabular}
	  \caption{Quantitative comparison results on unseen contaminated patterns across all available datasets. $\uparrow$ indicates the higher the better, and $\downarrow$ indicates the lower the better. \textbf{Bold} means the best results.}
	  \label{tab:contaminated_patterns}
\end{table*}

\subsection{Various Unseen Contaminated Patterns}
\begin{figure*}[!t]
\centering
\includegraphics[width=\linewidth]{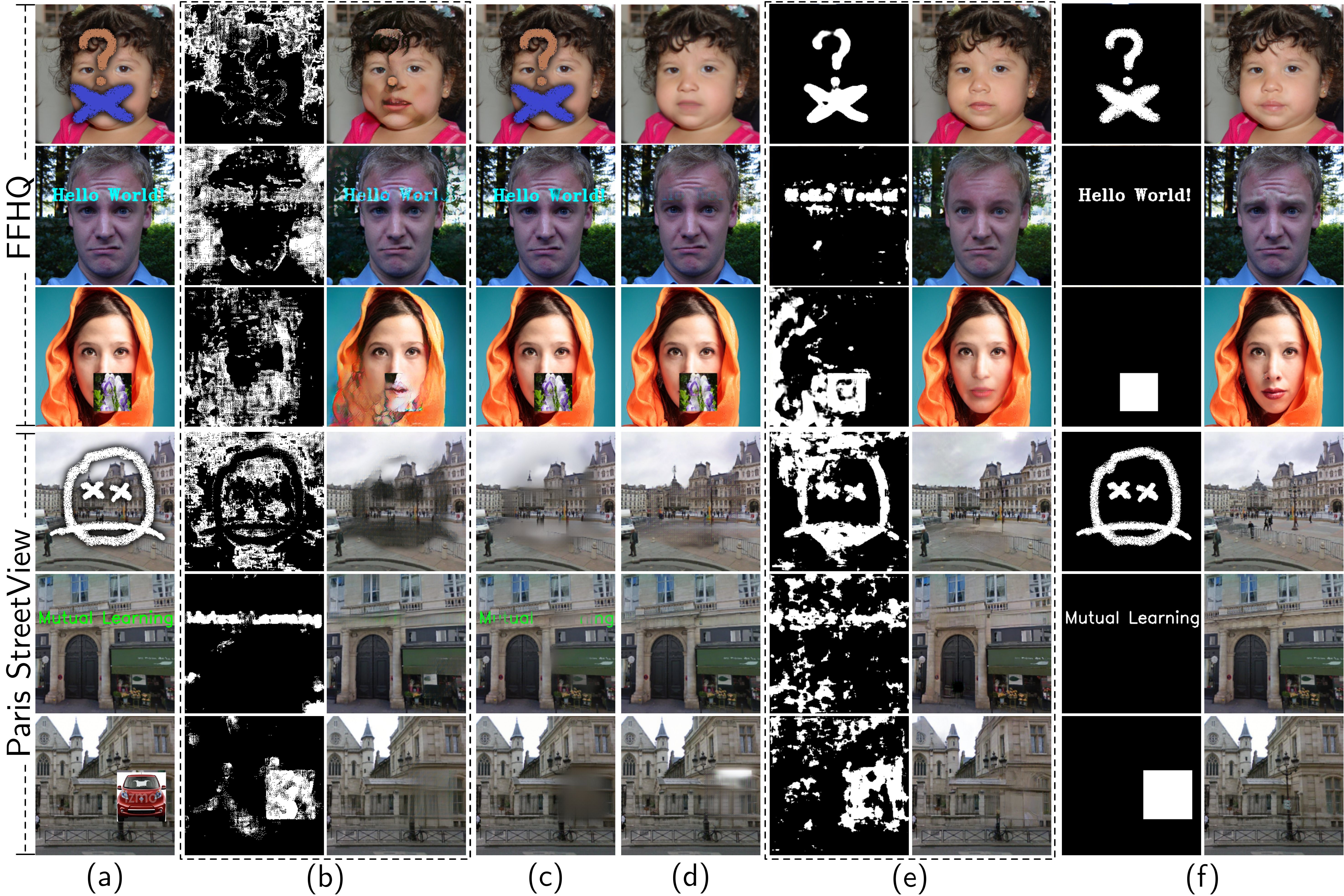}
\caption{Qualitative comparison results of unseen contaminated patterns for blind image inpainting on FFHQ and Paris StreetView: (a) Input contaminated images, (b) VCN (left: estimated masks, right: inpainted images), (c) OmniWavNet, (d) TransCNN-HAE, (e) our CAML (left: estimated masks, right: inpainted images), and (f) Ground Truth (left: masks, right: images).}
\label{fig:unseen}
\end{figure*}

We conduct quantitative and qualitative comparisons of our CAML with state-of-the-art blind image inpainting methods (VCN, TransCNN-HAE, and OmniWavNet) with various unseen contaminated patterns, such as graffiti (the contaminated ratio is 10-20\%), text (the contaminated ratio is 0-20\%), and image occlusion (the contaminated ratio is 0-20\%). VCN and our CAML are two-stage blind methods with mask estimation and image inpainting, while TransCNN-HAE and OmniWavNet are one-stage blind methods.
Table~\ref{tab:contaminated_patterns} and Figure~\ref{fig:unseen} demonstrate significant advantages of our CAML for dealing with various unseen contaminated patterns, proving its robustness against various unseen contamination. More qualitative comparison results of unseen contaminated patterns can be found at \url{https://github.com/zhenglab/CAML}.

\subsection{Real-World Cases}
\begin{figure}[!t]
\centering
\includegraphics[width=\linewidth]{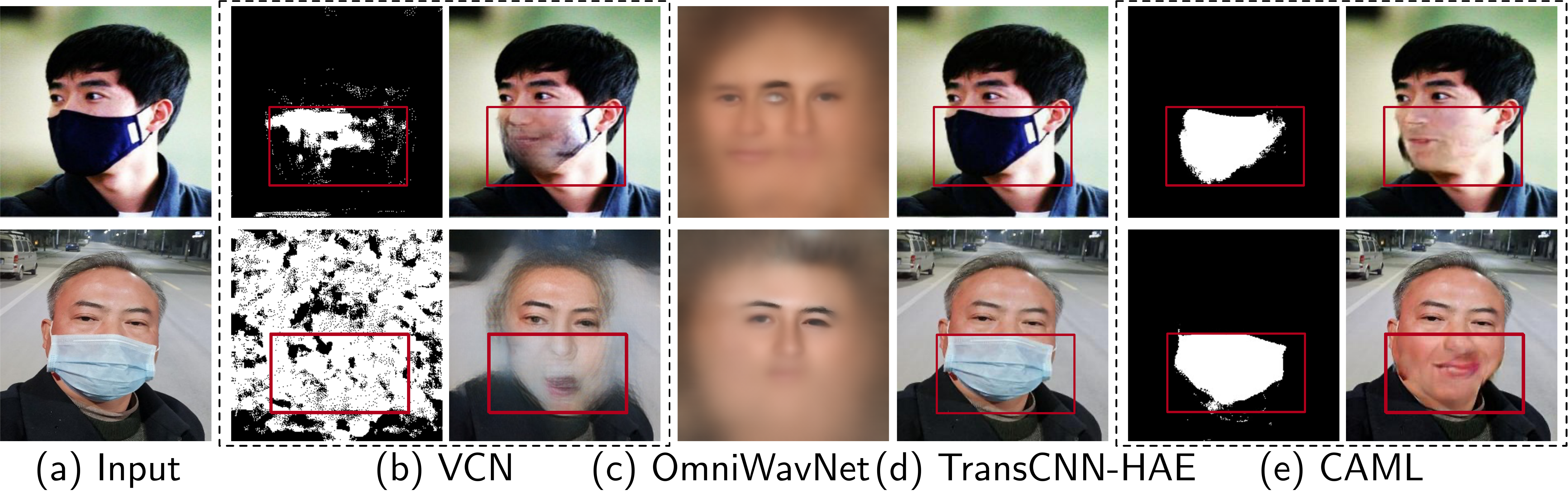}
\caption{Qualitative comparison results of real-world cases from Real-World Masked Face Dataset: (a) Input occluded faces, (b) VCN (left: estimated masks, right: inpainted images), (c) OmniWavNet, (d) TransCNN-HAE, and (e) our CAML (left: estimated masks, right: inpainted images).}
\label{fig:real_world}
\end{figure}

We further compare our CAML with blind image inpainting methods (VCN, TransCNN-HAE, and OmniWavNet) on real-world occluded faces from Real-World Masked Face Dataset~\citep{wang2020masked}. The real-world cases contain the real-world contaminated images and contaminated patterns.
Since real-world cases have no ground truth, we only conduct qualitative comparisons using different methods.  Our CAML, VCN, TransCNN-HAE, and OmniWavNet models are all trained on CelebA-HQ. Qualitative comparison results are shown in Figure~\ref{fig:real_world}, demonstrating that VCN cannot accurately estimate facemasks and restore reasonable and complete faces; OmniWavNet fails to identify facemasks and recover faces; TransCNN-HAE cannot identify occluded facemasks and remove them; our CAML can estimate occluded facemasks as accurately as possible and restore the main semantics of face, showing good generalization capability to real-world contamination.

\begin{table}[t!]
 	\setlength{\tabcolsep}{1mm}
 	\centering
 \footnotesize
		\begin{tabular}{c c c c c c c   c c }
			\hline
			\multirow{2}{*}{Number} & Embedding  &  \multirow{2}{*}{BCE$\downarrow$} & \multirow{2}{*}{PSNR$\uparrow$} & \multirow{2}{*}{SSIM$\uparrow$} & \multirow{2}{*}{FID$\downarrow$} & \multirow{2}{*}{LPIPS$\downarrow$} & Params$\downarrow$ & FLOPs$\downarrow$
       \\  
       &  Position &  & & & & & (M) & (G)
			\\
			\hline
			$0$ IGCM  & \multirow{2}{*}{-} &  \multirow{2}{*}{$0.591$} & \multirow{2}{*}{$20.45$}  & \multirow{2}{*}{$0.812$} & \multirow{2}{*}{$13.51$} & \multirow{2}{*}{$0.151$} & \multirow{2}{*}{$17.09$} & \multirow{2}{*}{$98.65$} 
			\\
                $0$ EGCM & & & & & & 
                \\
			\hline
			 \multirow{3}{*}{\shortstack {$1$ IGCM \\ $1$ EGCM}} &  Encoder$_{1st}$ & $0.577$ & $22.61$	& $0.862$ & $11.59$ & $0.134$ & $17.89$ & $125.93$\\
			 &  Encoder$_{2nd}$ & $0.575$ & $22.68$	& $0.864$ & $11.01$ & $0.132$ & $18.87$ & $125.86$  \\
    &  Encoder$_{3rd}$ & $0.574$ & $22.79$	& $0.868$ & $11.00$ & $0.135$   & $22.81$ & $125.80$ \\
		\hline
 		 \multirow{3}{*}{\shortstack {$2$ IGCM \\ $2$ EGCM}} &  Encoder$_{1st, 2nd}$ & $0.572$ & $22.91$	& $0.871$ & $9.80$ & $0.123$ & $19.20$ & $129.38$ \\
			 &  Encoder$_{1st, 3rd}$ & $0.570$ & $22.97$	& $0.871$ & $9.75$ & $0.123$ & $23.14$ & $129.31$ \\
    &  Encoder$_{2nd, 3rd}$ & $0.569$ & $23.11$	& $0.875$ & $9.61$ & $0.122$ & $24.12$ & $129.25$ \\
    \hline
    $3$ IGCM  &  \multirow{2}{*}{Encoder$_{1st, 2nd, 3rd}$} &  \multirow{2}{*}{$0.568$} &  \multirow{2}{*}{$23.40$}	& \multirow{2}{*}{$0.888$} & \multirow{2}{*}{$9.01$} & \multirow{2}{*}{$0.112$} & \multirow{2}{*}{$24.45$}  & \multirow{2}{*}{$132.76$}\\    
    $3$ EGCM  & & & & & &  \\ 
			\hline
    $4$ IGCM  &  \multirow{2}{*}{Encoder$_{1st, 2nd, 3rd}$} &  \multirow{2}{*}{$0.567$} &  \multirow{2}{*}{$23.47$}	& \multirow{2}{*}{$0.891$} & \multirow{2}{*}{$8.89$} & \multirow{2}{*}{$0.109$} & \multirow{2}{*}{$29.89$}  & \multirow{2}{*}{$136.83$}\\    
    $4$ EGCM  & & & & & &  \\ 
			\hline
    $5$ IGCM  &  \multirow{2}{*}{Encoder$_{1st, 2nd, 3rd}$} &  \multirow{2}{*}{$0.567$} &  \multirow{2}{*}{$23.46$}	& \multirow{2}{*}{$0.902$} & \multirow{2}{*}{$8.87$} & \multirow{2}{*}{$0.110$} & \multirow{2}{*}{$33.89$}  & \multirow{2}{*}{$145.05$}\\  
    $5$ EGCM  & & & & & &  \\ 
			\hline
    $6$ IGCM  &  \multirow{2}{*}{Encoder$_{1st, 2nd, 3rd}$} &  \multirow{2}{*}{$0.566$} &  \multirow{2}{*}{$23.50$}	& \multirow{2}{*}{$0.905$} & \multirow{2}{*}{$8.84$} & \multirow{2}{*}{$0.108$} & \multirow{2}{*}{$40.39$}  & \multirow{2}{*}{$153.27$}\\  
    $6$ EGCM  & & & & & &  \\ 
			\hline
    $7$ IGCM  &  \multirow{2}{*}{Encoder$_{1st, 2nd, 3rd}$} &  \multirow{2}{*}{$\mathbf{0.565}$} &  \multirow{2}{*}{$23.52$}	& \multirow{2}{*}{$\mathbf{0.906}$} & \multirow{2}{*}{$8.83$} & \multirow{2}{*}{$0.106$} & \multirow{2}{*}{$45.64$}  & \multirow{2}{*}{$161.49$}\\  
    $7$ EGCM  & & & & & &  \\ 
			\hline
    $8$ IGCM  &  \multirow{2}{*}{Encoder$_{1st, 2nd, 3rd}$} &  \multirow{2}{*}{$\mathbf{0.565}$} &  \multirow{2}{*}{$\mathbf{23.57}$}	& \multirow{2}{*}{$\mathbf{0.906}$} & \multirow{2}{*}{$\mathbf{8.81}$} & \multirow{2}{*}{$\mathbf{0.105}$} & \multirow{2}{*}{$50.89$}  & \multirow{2}{*}{$169.71$}\\    
    $8$ EGCM  & & & & & &  \\ 
			\hline
		\end{tabular}
    \caption{Ablation studies of the analysis of IGCM and EGCM learners with different numbers and embedding positions on FFHQ.}
\label{tab:EGCM_IGCM_ee}
\end{table}

\subsection{Ablation Studies of CAML}\label{sec:experiment:CAML}
We conduct ablation experiments to validate the efficacy of our CAML on FFHQ dataset with random masks, including the analysis of IGCM and EGCM learner, the efficacy of CMAT block's design and usage, the efficacy of CMAF block's design and usage, and the efficacy of iterative training. 

\textbf{Analysis of IGCM and EGCM Learner.}
We first analyze the impact of using different numbers of IGCM and EGCM learners in mask encoder and inpainting encoder separately on blind image inpainting within our CAML framework. We adjust the number of IGCM and EGCM learners (from $0$ to $8$) for the ablative experiment. Notably, we keep a consistent number of IGCM and EGCM learners to take full advantage of mutual learning.
Table~\ref{tab:EGCM_IGCM_ee} demonstrates that, our CAML with IGCM and EGCM learners significantly improves the performance of mask estimation and image inpainting. Moreover, more IGCM and EGCM learners are more helpful for mask estimation and image inpainting, and $8$ IGCM learners and $8$ EGCM learners perform best. However, when the number of IGCM and EGCM learners reaches $3$, the rate of performance improvement slows down and eventually saturates. This is primarily because, as the number of learners grows, deeper learners may begin to repeatedly process already captured complementary contextual information, limiting the model's ability to further enhance its representation. Additionally, too many learners also increase the complexity of the model, leading to higher computational costs and training difficulties. Therefore, considering the balance between performance improvement and computational resources, we ultimately chose to use 3 IGCM and 3 EGCM learners in our method for capturing more complementary contextual information and fully exploiting the benefits of mutual learning.


We then conduct the ablative experiment to study the role of the embedding position of IGCM and EGCM learners in mask encoder and inpainting encoder respectively on blind image inpainting within our CAML framework. Specifically, we consider all the different embedding positions of IGCM and EGCM learners as follows: embedding IGCM and EGCM learners separately into mask and inpainting encoder after (1) the first layer (Encoder$_{1st}$), (2) the second layer (Encoder$_{2nd}$), (3) the third layer (Encoder$_{3rd}$), (4) the first and second layers (Encoder$_{1st,2nd}$), (5) the first and third layers (Encoder$_{1st,3rd}$), (6) the second and third layers (Encoder$_{2nd,3rd}$), as well as (7) the first, second, and third layers (Encoder$_{1st,2nd,3rd}$). Table~\ref{tab:EGCM_IGCM_ee} shows that, under the same number of IGCM and EGCM learners, IGCM and EGCM learners embedded in deep layers play a more important role in mask estimation and image inpainting than embedded in shallow layers. This is because the mask and inpainting encoder have abundant contextual semantics and details respectively in the deeper layer. However, increasing the number of each learner can significantly improve the performance more than embedding it in deeper layers. Therefore, we select the embedding positions of the IGCM and EGCM leaners as the first, second, and third layers of the mask encoder and inpainting encoder respectively for our CAML framework.

\begin{figure}[!t]
\centering
\includegraphics[width=\linewidth]{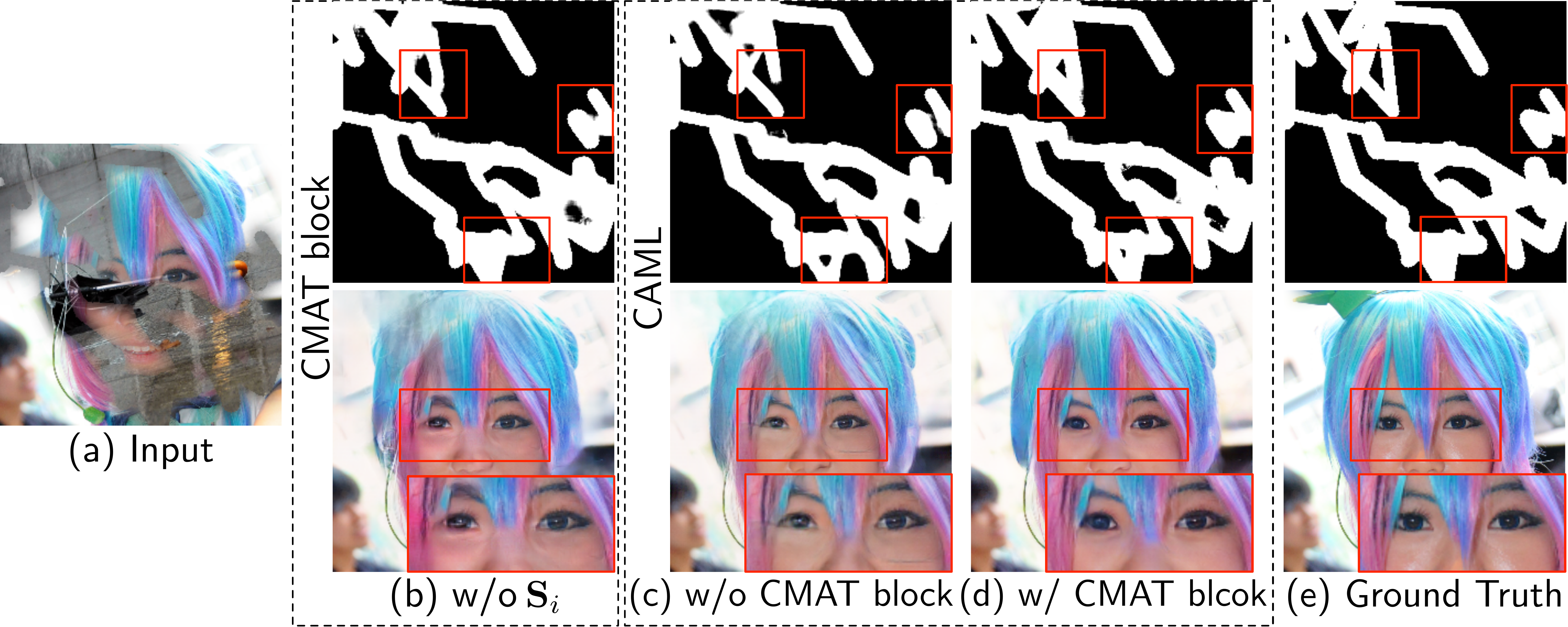}
\caption{Visual examples for ablation study of validating the efficacy of CMAT block's design and usage on FFHQ: (a) Input contaminated image, CMAT block (b) w/o $\mathbf{S}_i$ (top: estimated mask, bottom: inpainted image), CAML (c) w/o CMAT block and (d) our w/ CMAT block (top: estimated mask, bottom: inpainted image), and (e) Ground Truth (top: mask, bottom: image).}
\label{fig:exp:CMAT}
\end{figure}

\textbf{Efficacy of CMAT Block's Design and Usage.} \label{sec:exp-aib}
We next conduct ablation study to validate the efficacy of CMAT block's design and usage: for design, (1) CMAT block without soft gatings $\mathbf{S}_{i}$ (w/o $\mathbf{S}_i$)~\citep{wu2019mutual} that the guiding features $f_{G_i}$ are directly used to generate the context-aware parameters $\psi_{i}$; for usage, CAML (2) without CMAT block (w/o CMAT block) and (3) with CMAT block (w/ CMAT block).
Table~\ref{tab:CMAT_ee} indicates the efficacy of our CMAT block's design (our CAML w/ CMAT block performs better than CMAT block w/o $\mathbf{S}_i$) and CMAT block's usage  (our CAML w/ CMAT block outperforms w/o CMAT block). Figure~\ref{fig:exp:CMAT} demonstrates that CMAT block w/o $\mathbf{S}_i$ and CAML w/o CMAT block produce the inaccurate mask as well as generate the image with unreasonable contents (eyes), while our CAML w/ CMAT block estimates the more accurate mask and synthesizes the more realistic image.

\begin{table}[t!]
\footnotesize
 	\setlength{\tabcolsep}{2mm}
 	\centering
		\begin{tabular}{c c c c c c c c c}
			\hline
			\multicolumn{2}{c}{Method} && BCE$\downarrow$ & PSNR$\uparrow$ & SSIM$\uparrow$ & FID$\downarrow$ & LPIPS$\downarrow$
			\\
\cline{1-2} \cline{4-8}
			CMAT block & w/o $\mathbf{S}_i$ && $0.572$ & $22.67$ & $0.861$ & $10.68$ & $0.130$
			\\
			\hline
			 \multirow{2}{*}{CAML} &  w/o CMAT block && $0.591$ & $21.78$	& $0.814$ & $12.12$ & $0.147$\\
			& w/ CMAT block && $\mathbf{0.568}$ & $\mathbf{23.40}$ & $\mathbf{0.888}$ & $\mathbf{9.01}$ & $\mathbf{0.112}$ 
			\\
			\hline
		\end{tabular}
	\caption{Ablation studies of validating the efficacy of CMAT block's design and usage on FFHQ.}
	\label{tab:CMAT_ee}
\end{table}

\begin{table}[t]
\footnotesize

 	\setlength{\tabcolsep}{2mm}
 	\centering
		\begin{tabular}{c c  c c c c c c}
			\hline
		 \multicolumn{2}{c}{Method} & & PSNR$\uparrow$ & SSIM$\uparrow$ & FID$\downarrow$ & LPIPS$\downarrow$
			\\
\cline{1-2} \cline{4-7}
			\multirow{5}{*}{CMAF block}& 
			w/o $f^{c}_{I_i}$ & & $22.84$ & $0.871$ & $10.05$ & $0.141$
			\\
			& w/o $f^{nc}_{I_i}$ & & $20.13$ & $0.718$ & $25.22$  & $0.275$
			\\ 
               & Min && $9.30$ & $0.029$ & $214.58$ & $0.743$\\
           & Max && $22.90$ & $0.875$ & $10.90$ & $0.125$ \\
           & Average && $21.76$ & $0.860$ & $12.04$ & $0.141$ \\
		\hline
			\multirow{2}{*}{CAML} & w/o CMAF block & & $22.06$ & $0.865$ & $10.84$ 
            & $0.133$\\
			& w/ CMAF block & & $\mathbf{23.40}$ & $\mathbf{0.888}$ & $\mathbf{9.01}$ & $\mathbf{0.112}$
			\\
	\hline
		\end{tabular}
	\caption{Ablation studies of validating the efficacy of CMAF block's design and usage on FFHQ.}
	\label{tab:CMAF}
\end{table}

\begin{figure}[t]
\centering
\includegraphics[width=\linewidth]{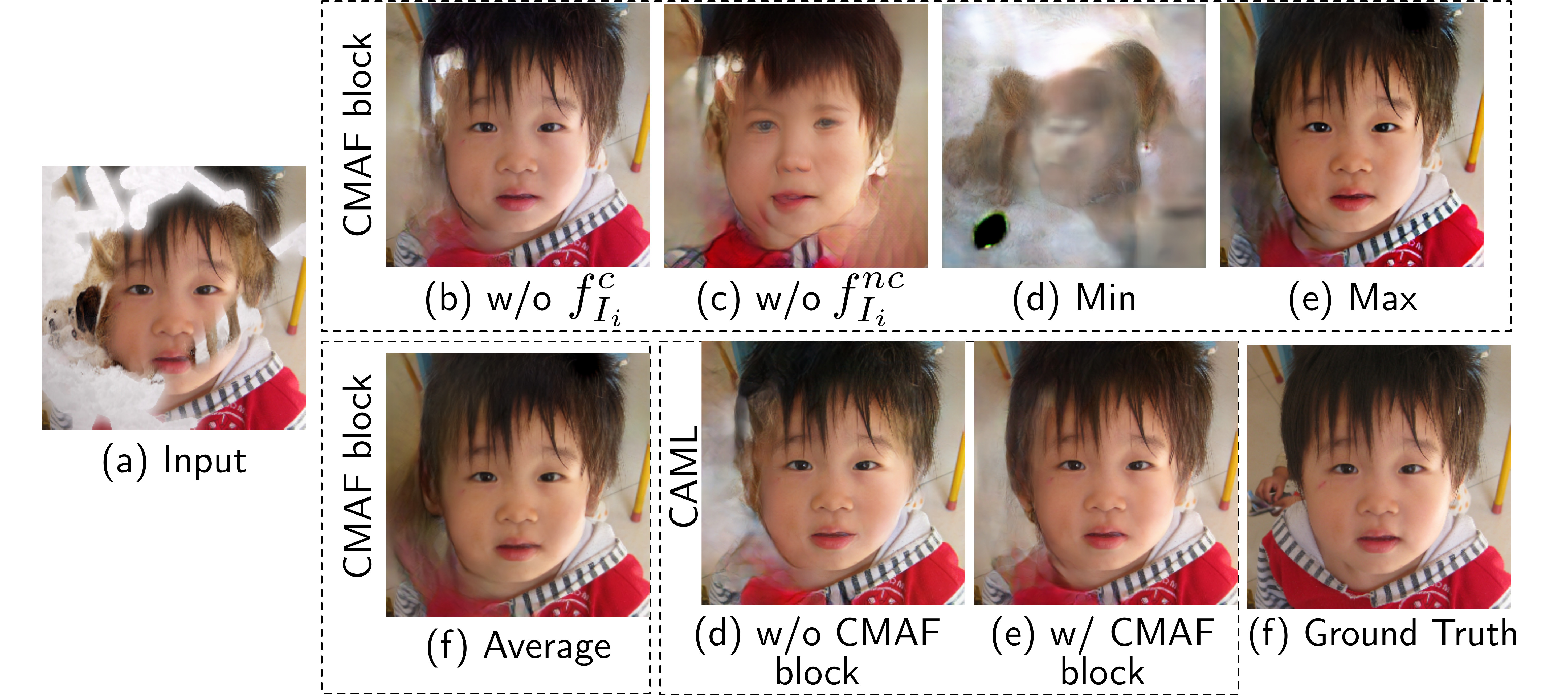}
\caption{Visual examples for ablation study of validating the efficacy of CMAF block's design and usage on FFHQ.}
\label{fig:exp:CMAF}
\end{figure}

\textbf{Efficacy of CMAF Block's Design and Usage.}
We further conduct ablation study to validate the efficacy of CMAF block's design and usage: for design, CMAF block (1) without contaminated regions (w/o $f^c_{I_i}$) and (2) without non-contaminated regions (w/o $f^{nc}_{I_i}$), contaminated and non-contaminated regions in CMAF block with different fusion ways (3) minimum fusion (Min), (4) maximum fusion (Max), and (5) average fusion (Average); for usage, CAML (6) without CMAF block (w/o CMAF block) and (7) with CMAF block (w/ CMAF block). 
Table~\ref{tab:CMAF} indicates the superiority of CMAF block in design and usage for image inpainting. Figure~\ref{fig:exp:CMAF} also demonstrates that CAML w/o CMAF block or CMAF block w/o $f^{c}_{I_i}$ generates image with feature of contaminated region, CMAF block w/o $f^{nc}_{I_i}$ leads to blurry result due to inadequate learning of non-contaminated region, maximum fusion (Max) may lead to the loss of valuable information, average fusion (Average) can dilute the information, resulting in unreasonable outcomes, and minimum fusion (Min) may even fail to generate coherent semantics, while our CAML w/ CMAF block with element-wise product adaptive fusion yields the best results.

\textbf{Efficacy of Iterative Training.} \label{sec:exp-iterative}
We lastly validate the efficacy of the iterative training strategy for our CAML with simultaneous training mask estimation and image inpainting (Simultaneous) as well as iterative training for ablation study.
For simultaneous training, we also could consider a multi-task approach~\cite{khattar2021cross} where both mask estimation and image inpainting tasks share a common encoder (Multi-task). 
Iterative training can be divided into two cases: image inpainting first (II$_{first}$) and mask estimation first (ME$_{first}$).
Table~\ref{tab:training strategy} shows that, iterative training is superior to simultaneous training. The performance of CAML with simultaneous or iterative training is significantly better than that of the multi-task architecture with a shared encoder during simultaneous training. This is because directly employing a multi-task approach with a shared common encoder would constrain the specificity of the two tasks, leading to feature confusion and ultimately hindering their mutual learning. Mask estimation is first trained (ME$_{first}$) to exhibit the best performance. Considering that, during training, image inpainting requires both contextual features and the estimated mask generated by mask estimation, while mask estimation requires only the contextual features obtained in image inpainting. The performance of image inpainting is more dependent on the quality of mask estimation. Therefore, mask estimation is trained first to improve its performance, which will be more conducive to the improvement of image inpainting performance.

\begin{table}[!h]

 \footnotesize
 	\setlength{\tabcolsep}{2.5mm}
 	\centering
		\begin{tabular}{c c c c c c c}
			\hline
			Method  & BCE$\downarrow$ & PSNR$\uparrow$ & SSIM$\uparrow$ & FID$\downarrow$ & LPIPS$\downarrow$
			\\ 
			\hline
                Multi-task & $0.702$ & $21.73$ & $0.845$ & $11.74$ & $0.147$ \\
			Simultaneous  & $0.570$ & $23.21$ & $0.885$ & $9.42$ &  $0.117$
			\\
			II$_{first}$ & $\mathbf{0.567}$ & $23.23$ & $0.884$ & $9.34$ &  $0.118$
			\\
			ME$_{first}$ & $0.568$ & $\mathbf{23.40}$ & $\mathbf{0.888}$ & $\mathbf{9.01}$ & $\mathbf{0.112}$
			\\
			\hline
		\end{tabular}
	\caption{Ablation studies of validating the efficacy of iterative training on FFHQ.}
	\label{tab:training strategy}
\end{table}

\section{Beyond the Blind Image Inpainting}
In this section, we evaluate the proposed CAML in snow removal, shadow removal, and watermark removal tasks beyond the blind image inpainting. We retrain CAML on the corresponding dataset for the snow/shadow/watermark removal task, and compare it with previous state-of-the-art as well as recent snow/shadow/watermark removal methods. For fair comparisons, the results produced by these methods are generated using the original code released by the corresponding authors or directly provided by them.

\subsection{Snow Removal}
\begin{table}[t]

  \footnotesize
 	\setlength{\tabcolsep}{2mm}
 	\centering
		\begin{tabular}{c c c c c c c c}
			\hline
			\multicolumn{1}{c}{\multirow{2}{*}{Method}}  & \multicolumn{3}{c}{Snow100K} && \multicolumn{3}{c}{CSD} \\
   \cline{2-4} \cline{6-8}
			& PSNR$\uparrow$ & SSIM$\uparrow$ &  
   LPIPS$\downarrow$ && PSNR$\uparrow$ & SSIM$\uparrow$ & LPIPS$\downarrow$
			\\ 
			\hline
			DesnowNet &  $23.12$ & $0.780$ & $0.25$	&& $20.63$ & $0.777$ & $0.30$		\\
	
		HDCWNet & $21.29$ & $0.640$ & $0.40$ && $28.66$ & $0.892$ & $0.11$ \\

   DDMSNet & $29.05$ & $0.891$ & $0.10$ && $24.97$ & $0.905$ & $0.08$ \\

      TKL & $30.50$ & $0.890$ & $0.09$ && $33.35$ & $0.954$ & $0.03$ \\

      CAML & $\mathbf{31.30}$ & $\mathbf{0.973}$ & $\mathbf{0.06}$ && $\mathbf{34.03}$ & $\mathbf{0.984}$ & $\mathbf{0.01}$\\
			\hline
		\end{tabular}
    \vspace*{-2pt}
	\caption{Quantitative comparison of our CAML with existing state-of-the-art snow removal methods (DesnowNet, HDCWNet, DDMSNet, and TKL) on Snow100K and CSD.}
	\label{tab:image_snow_removal}
\end{table}

\begin{figure}[t]
\centering
\includegraphics[width=\linewidth]{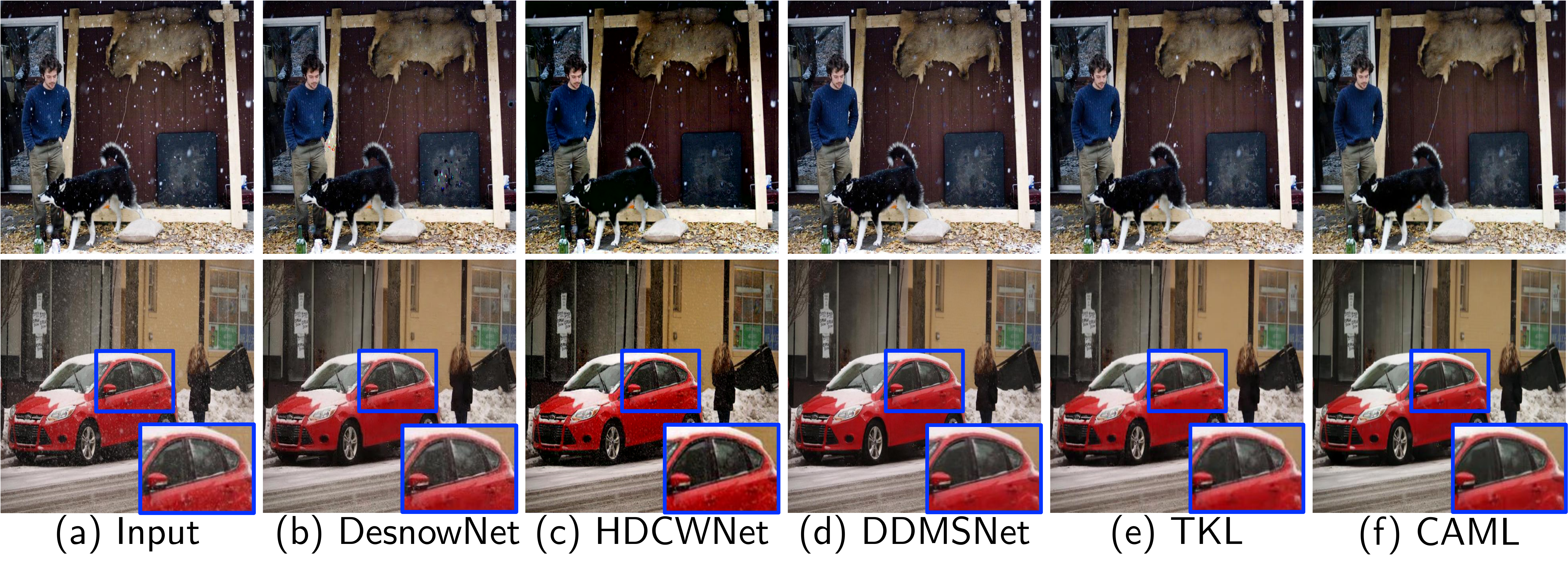}
\caption{Qualitative comparison results of snow removal on the real-world dataset: (a) Input snowy images, (b) DesnowNet, (c) HDCWNet, (d) DDMSNet, (e) TKL, and (f) our CAML.}
\label{fig:image_snow_removal}
\end{figure}

Beyond the blind image inpainting, we first employ our CAML framework for snow removal task, where we decompose this task into two mutual sub-tasks, \ie, snow estimation and snow removal. Compared with some atmospheric phenomena (\eg, rainy, haze), due to the non-transparency property as well as complex shapes and sizes of snow, snow removal is more challenging. In this experiment, we compare our CAML with state-of-the-art snow removal methods DesnowNet~\cite{liu2018desnownet}, HDCWNet~\cite{chen2021all}, DDMSNet~\cite{zhang2021deep}, and TKL~\cite{chen2022learning} on the synthetic datasets: Snow100K~\cite{liu2018desnownet} and CSD~\cite{chen2021all} as well as the real-world dataset~\cite{liu2018desnownet}. 
Following~\cite{liu2018desnownet,chen2021all}, we adopt PSNR, SSIM, and LPIPS to evaluate model performance. The quantitative comparison results are presented in Table~\ref{tab:image_snow_removal}, indicating that our CAML outperforms DesnowNet, HDCWNet, DDMSNet, and TKL. Figure~\ref{fig:image_snow_removal} shows the qualitative comparison results on the real-world dataset, demonstrating that our CAML can remove snow well and has better generalization ability in real-world scenarios. Thanks to the mutual learning of snow estimation and snow removal, our CAML can estimate complex snow types and achieve superior snow removal performance.

\begin{table}[!t]
 	\setlength{\tabcolsep}{1mm}
    \footnotesize
 	\centering
		\begin{tabular}{c c c c c c c c}
				\hline
			Method  & \multicolumn{3}{c}{ISTD} && \multicolumn{3}{c}{SRD} \\
   \cline{2-4} \cline{6-8}
			(RMSE$\downarrow$) & Shadow & Non-Shadow & All && Shadow & Non-Shadow & All
			\\ 
			\hline
			DSC & $9.76$ & $6.14$ & $6.67$ && $10.89$ & $4.99$ & $6.23$	 	\\

			DHAN & $8.14$ & $6.04$ & $6.37$ && $8.94$ & $4.80$ & $5.67$   \\

		AEF & $7.77$ & $5.56$ & $5.92$ && $8.56$ &$5.75$ & $6.61$ \\

        ShadowFormer & $6.76$ & $4.44$ & $4.79$ && $5.90$  & $3.44$ & $4.04$\\

        CAML  & $\mathbf{5.59}$ & $\mathbf{4.36}$ & $\mathbf{4.57}$ && $\mathbf{5.16}$ & $\mathbf{2.67}$ & $\mathbf{3.88}$
			\\
			\hline
		\end{tabular}
	\caption{Quantitative comparison of our CAML with existing state-of-the-art shadow removal methods (DSC, DHAN, AEF, and ShadowFormer) on ISTD and SRD. }
	\label{tab:image_shadow_removal}
\end{table}

\begin{figure}[!t]
\centering
\includegraphics[width=\linewidth]{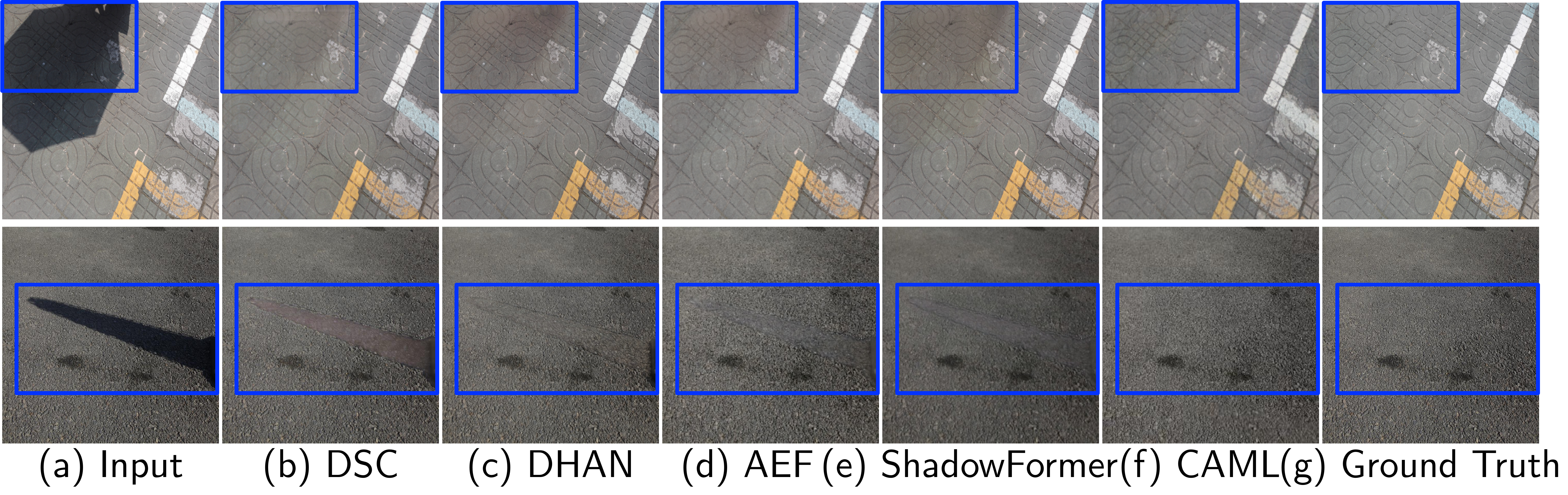}
\caption{Qualitative comparison results of shadow removal: (a) Input shadowed images, (b) DSC, (c) DHAN, (d) AEF, (e) ShadowFormer, (f) our CAML, and (g) Ground Truth. Top: ISTD dataset. Bottom: SRD dataset.}
\label{fig:image_shadow_removal}
\end{figure}

\subsection{Shadow Removal}

We then apply our CAML framework to the shadow removal task, where we also decompose this task into two mutual sub-tasks, \ie, shadow estimation and shadow removal. Shadows usually form when light source is blocked, also variant color and illumination distortion in shadows may degrade the model performance on many computer vision tasks, such as object detection and semantic segmentation, thus removing these shadows is beneficial but full of challenges. We compare our CAML with state-of-the-art shadow removal methods DSC~\citep{hu2019direction}, DHAN~\citep{cun2020towards}, AEF~\citep{fu2021auto}, and ShadowFormer~\citep{guo2023shadowformer} on ISTD~\citep{wang2018stacked} and SRD~\citep{qu2017deshadownet} datasets. Following~\citep{hu2019direction,cun2020towards,fu2021auto}, we use root mean square error (RMSE) in LAB color space to evaluate the shadow removal performance in shadow, non-shadow, and the whole regions. Table~\ref{tab:image_shadow_removal} and Figure~\ref{fig:image_shadow_removal} show quantitative and qualitative comparison results respectively, indicating that our CAML can estimate shadow and obtain traceless results of shadow removal, which benefits from the mutual learning of shadow estimation and shadow removal.

\begin{table}[!t]
 	\setlength{\tabcolsep}{2mm}
      \footnotesize
 	\centering
		\begin{tabular}{c c c c c c c c}
			\hline
			\multicolumn{1}{c}{\multirow{2}{*}{Method}}  & \multicolumn{3}{c}{LOGO30k} && \multicolumn{3}{c}{CLWD} \\
   \cline{2-4} \cline{6-8}
			& PSNR$\uparrow$ & SSIM$\uparrow$ & RMSE$\downarrow$ && PSNR$\uparrow$ & SSIM$\uparrow$ & RMSE$\downarrow$
			\\
	\hline
	BVMR & $38.28$ & $0.985$ & $3.87$ && $35.89$ & $0.973$ & $5.02$\\

	SplitNet & $41.27$ & $0.991$ & $2.74$ && $37.41$ & $0.978$ & $4.23$  \\

	SLBA  & $41.09$ & $0.995$  & $3.39$ && $38.28$ & $0.981$
    & $3.76$\\

	CAML & $\mathbf{42.13}$ & $\mathbf{0.996}$ & $\mathbf{2.70}$ && $\mathbf{38.32}$ & $\mathbf{0.993}$ & $\mathbf{2.52}$\\
				\hline
		\end{tabular}
	\caption{Quantitative comparison of our CAML with existing state-of-the-art watermark removal methods (BVMR, SplitNet, and SLBA) on LOGO30k and CLWD.}
	\label{tab:watermark_removal}
\end{table}

\begin{figure}[!t]
\centering
\includegraphics[width=\linewidth]{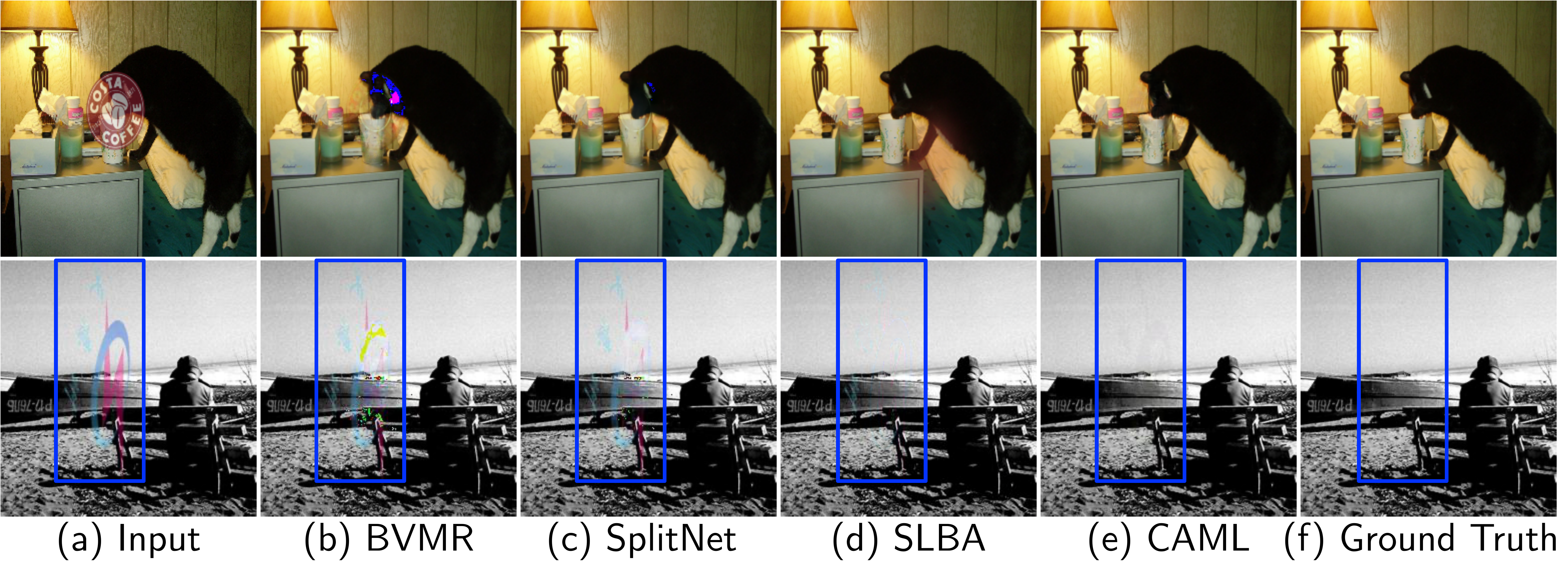}
\caption{Qualitative comparison results of watermark removal: (a) Input watermarked images, (b) BVMR, (c) SplitNet, (d) SLBA, (e) our CAML, and (f) Ground Truth. Top: LOGO30K dataset. Bottom: CLWD dataset.}
\label{fig:image_watermark_removal}
\end{figure}

\subsection{Watermark Removal}

We finally employ our CAML framework to watermark removal task, where we similarly decompose watermark removal into two mutual sub-tasks, \ie, watermark estimation and watermark removal. Watermarks usually contain complex patterns, such as warped symbols, thin lines, shadow effects, and have diverse sizes, shapes, and colors. Watermark removal aims to remove watermarks from watermarked images and restore the content of images. We compare our CAML with state-of-the-art watermark removal methods BVMR~\citep{hertz2019blind}, SplitNet~\citep{cun2021split}, and SLBA~\citep{liang2021visible} on LOGO30K~\citep{cun2021split} and CLWD~\citep{liu2021wdnet} datasets. Following~\citep{liang2021visible}, we adopt PSNR, SSIM, and RMSE distance to evaluate model performance. Table~\ref{tab:watermark_removal} and Figure~\ref{fig:image_watermark_removal} demonstrate that our CAML can estimate and remove watermarks as well as recover realistic content. We argue that the advantage of CAML mainly stems from mutual learning, which improves the representation ability of both watermark estimation and watermark removal.

\section{Limitation and Future Work}
In this work, we demonstrate that context-aware mutual learning between mask estimation and image inpainting is advantageous for enhancing the performance of blind image inpainting. However, our approach still has a limitation. The additional design of mutual learning between mask estimation and image inpainting does result in higher computational costs for our model. In our future work, we intend to investigate more lightweight two-stage mutual learning model.

\section{Conclusion}
In this work, we propose a novel CAML framework for blind image inpainting. The framework aims to leverage context-aware mutual learning between mask estimation and image inpainting to boost the performance of blind image inpainting. Specifically, our framework includes two novel learners: Inpainting-Guided Context-Mutual (IGCM) learner and Estimation-Guided Context-Mutual (EGCM) learner. IGCM learner is utilized to acquire supplementary contextual details from image inpainting for assisting mask estimation, while EGCM learner is employed to capture complementary contextual semantics from mask estimation for enhancing image inpainting. Moreover, CAML framework facilitates an iterative interaction between mask estimation and image inpainting to guide each other for fully absorbing contextual information. 
Extensive experiments demonstrate that our CAML achieves state-of-the-art performance on various datasets with complexity contaminations for blind image inpainting. 
Ablation studies further validate the efficacy of our CAML framework.
Furthermore, we employ our CAML on additional vision tasks beyond blind image inpainting, \ie, snow removal, shadow removal, and watermark removal, further illustrating the superiority of our work. We hope that our work opens up new avenues for blind image inpainting and its applications.

  \section*{Acknowledgments}

This work was supported by the National Natural Science Foundation of China (No. 62171421) and the TaiShan Scholars Youth Expert Program of Shandong Province (No. tsqn202306096).

\bibliographystyle{elsarticle-num}
 \bibliography{egbib}






\end{document}